\pdfoutput=1

\documentclass[11pt]{article}

\usepackage[preprint]{acl}

\usepackage{times}
\usepackage{latexsym}
\usepackage[T1]{fontenc}
\usepackage[utf8]{inputenc}
\usepackage{microtype}
\usepackage{inconsolata}
\usepackage{graphicx}

\usepackage{float} 
\usepackage{graphicx} 
\usepackage{subcaption} 
\usepackage{multirow} 
\usepackage{diagbox} 
\usepackage{color,soul} 
\usepackage{linguex} 
\usepackage{amsmath}
\usepackage{bm} 
\usepackage{arydshln} 
\usepackage{comment} 
\usepackage{xfrac} 

\usepackage{tikz}
\usepackage{scalerel}
\usepackage{pict2e}
\usepackage{tkz-euclide}
\usetikzlibrary{calc}
\usetikzlibrary{patterns,arrows.meta}
\usetikzlibrary{shadows}
\usetikzlibrary{external}
\usepackage{adjustbox}

\definecolor{primaryLight}{HTML}{C0392B}
\definecolor{new_green}{rgb}{0.35, 0.69, 0.19}
\definecolor{new_red}{rgb}{0.85, 0, 0.}
\definecolor{new_blue}{rgb}{0., 0.5, 1.}
\definecolor{new_yellow}{rgb}{0.98, 0.85, 0.37}
\definecolor{new_gray}{rgb}{0.86, 0.86, 0.86}
\definecolor{new_beige_light}{rgb}{0.98, 0.92, 0.84}
\definecolor{new_beige_dark}{rgb}{0.8, 0.58, 0.46}

\title{A Multi-Task and Multi-Label Classification Model \\ for Implicit Discourse Relation Recognition}

\author{Nelson Filipe Costa \and Leila Kosseim \\
        Computational Linguistics at Concordia (CLaC) Laboratory \\
        Department of Computer Science and Software Engineering \\
        Concordia University, Montr\'eal, Qu\'ebec, Canada \\
        \texttt{nelsonfilipe.costa@mail.concordia.ca} \\
        \texttt{leila.kosseim@concordia.ca}}

\begin{document}

\maketitle


\begin{abstract}

We propose a novel multi-label classification approach to implicit discourse relation recognition (IDRR). Our approach features a multi-task model that jointly learns multi-label representations of implicit discourse relations across all three sense levels in the PDTB~3.0 framework. The model can also be adapted to the traditional single-label IDRR setting by selecting the sense with the highest probability in the multi-label representation. We conduct extensive experiments to identify optimal model configurations and loss functions in both settings. Our approach establishes the first benchmark for multi-label IDRR and achieves SOTA results on single-label IDRR using DiscoGeM. Finally, we evaluate our model on the PDTB~3.0 corpus in the single-label setting, presenting the first analysis of transfer learning between the DiscoGeM and PDTB~3.0 corpora for IDRR.

\end{abstract}


\section{Introduction}

Implicit discourse relation recognition (IDRR) is one of the most challenging tasks in computational discourse analysis. Its goal is to identify the sense of discourse relations connecting text arguments in the absence of explicit discourse connectives, such as \textit{but} and \textit{because}. The Penn Discourse Treebank (PDTB)~\citep{miltsakaki2004penn,prasad2008penn} organizes discourse senses across three hierarchical levels with increasing degrees of detail~\citep{webber2019penn}. However, understanding the sense of a discourse relation can be a complex and subjective task without the guidance of discourse connectives. Consider the two arguments of the implicit discourse relation below:

\textbf{Arg1:} The lights flickered.

\textbf{Arg2:} The power went out.

In this example, it is not clear whether the two arguments of the relation are related by a temporal or causality relation. One could interpret that the lights flicked and \textit{then} the power went out, or that the lights flickered \textit{because} the power went out. This ambiguity illustrates the biggest challenge of IDRR. Despite its difficulty, IDRR plays a crucial role in downstream tasks that rely on text coherence. In summarization, for example, it helps preserve the logical flow and rhetorical intent of the source text. In dialogue systems, it supports coherence and intent modeling by inferring how utterances relate to each other, improving dialogue discourse parsing \citep{li-etal-2024-dialogue,thompson-etal-2024-llamipa}.

To date, most research on IDRR has relied on the different iterations of the PDTB corpus~\citep{corpus-pdtb-2,corpus-pdtb-3}. However, despite significant efforts, SOTA performance on IDRR has plateaued in recent years at F1-Scores of $71.59$ at level-1 and $57.62$ at level-2~\citep{long-webber-2022-facilitating,liu-strube-2023-annotation,chan-etal-2023-DiscoPrompt,zhao-etal-2023-infusing,zeng-etal-2024-global,long2024leveraging}. One possible reason may be the inherently subjective nature of implicit discourse interpretation, which can be difficult to capture using predominantly single-label annotated corpora~\citep{stede2008disambiguation,rohde-etal-2016-filling,scholman2017examples,hoek-etal-2021-less}. The idea of possible multiple relations between discourse arguments had already been considered in the Segmented Discourse Representation Theory (SDRT) \citep{asher2003logics} and the ambiguity in the annotation of implicit relations has been further highlighted by the challenges in mapping them across discourse frameworks~\citep{demberg2019compatible,costa2023mapping}. In light of these challenges, recent studies have advocated for the multi-label annotation of implicit relations to better capture their nuanced and complex nature in discourse corpora~\citep{yung-etal-2019-crowdsourcing,pyatkin-etal-2020-qadiscourse,scholman-etal-2022-discogem,scholman-etal-2022-design,pyatkin2023design,yung-demberg-2025-crowdsourcing}. This shift in perspective has led to the creation of DiscoGeM~\citep{scholman-etal-2022-discogem} - the first multi-label annotated corpus of implicit discourse relations.

Following recent work in NLP~\citep{pavlick-kwiatkowski-2019-inherent,nie-etal-2020-learn,basile-etal-2021-need,fornaciari-etal-2021-beyond,uma2021learning,plank-2022-problem,jiang-marneffe-2022-investigating}, we see disagreement between humans annotators not as noise, but as an opportunity to enhance the relevance of IDRR models by capturing the complexity of human interpretations. This paper addresses the subjectivity in IDRR by proposing a multi-task classification model capable of learning both multi-label and single-label representations of implicit discourse relations. Our model is trained on the DiscoGeM corpus and uses a multi-task architecture to leverage the interdependence between senses and jointly learn sense representations across all three sense levels in the PDTB~3.0 hierarchy. The main contributions of this work are:

\begin{itemize}
    \item We present the first multi-label approach in IDRR which produces probability distributions over all possible sense labels of a discourse relation using the DiscoGeM corpus. Our multi-label model can then be adapted to the traditional single-label task by selecting the label with the highest probability.

    \item We conduct extensive experiments, comparing different pre-trained language models as encoders of our model and evaluating how different loss functions impact performance in multi-label and single-label IDRR.

    \item We establish the first benchmark on multi-label IDRR and achieve SOTA results on single-label IDRR using DiscoGeM.

    \item We present an in-depth analysis on the potential of transfer learning between the DiscoGeM and the PDTB~3.0 corpora on single-label IDRR.
\end{itemize}





\section{Previous Work}

Despite the recent advances in natural language understanding, IDRR remains one of the most challenging tasks in discourse analysis. Most previous work addressed IDRR as a single-label classification task by fine-tuning~\citep{long-webber-2022-facilitating,liu-strube-2023-annotation} or prompt-tuning~\citep{chan-etal-2023-DiscoPrompt,zhao-etal-2023-infusing,zeng-etal-2024-global,long2024leveraging} pre-trained language models (PLMs). More precisely, \citet{long-webber-2022-facilitating} apply contrastive learning and augment training data with explicit connectives from PDTB~3.0 metadata to fine-tune their model, while \citet{liu-strube-2023-annotation} propose a two-step pipeline that first generates an explicit connective for each relation to fine-tune a classifier on the modified input. \citet{chan-etal-2023-DiscoPrompt} inject hierarchical structure and connective-based explanations into prompts, enabling joint predictions across all three PDTB sense levels. \citet{zhao-etal-2023-infusing} address data scarcity in IDRR by using a parameter-efficient prompt-tuning framework that incorporates hierarchical label guidance into the verbalizer. Building on this, \citet{zeng-etal-2024-global} propose a prompt-tuning approach that integrates both global and local hierarchical label information into the verbalizer to improve output alignment with pre-trained objectives. Finally, \citet{long2024leveraging} introduce a prototype-based verbalizer informed by the PDTB~3.0 sense hierarchy, combining contrastive and prototype learning to eliminate the need for manually designed verbalizers. While most of these approaches~\citep{long-webber-2022-facilitating,liu-strube-2023-annotation,zhao-etal-2023-infusing,zeng-etal-2024-global,long2024leveraging} use RoBERTa$_{\text{base}}$~\citep{liu2019roberta} as their PLM, \citet{chan-etal-2023-DiscoPrompt} uses T5$_{\text{base}}$~\citep{raffel2020exploring}. Another line of work tried to solve single-label IDRR by directly prompting large language models (LLMs) through prompt-engineering~\citep{chan-etal-2024-exploring,yung-etal-2024-prompting}. However, both works show that the results obtained through directly prompting LLMs in zero-shot and few-shot settings are still far behind from those obtained through fine-tuning and prompt-tuning PLMs.

The shift toward multi-label annotation has stirred an initial wave of research in multi-label IDRR. For instance, \citet{long2024multi} used the 4.9\% of implicit relations in the PDTB~3.0 corpus that are annotated with two senses to build a model capable of predicting up to two senses per instance. However, given that the 95.1\% of PDTB~3.0 annotations remain single-label, their model predominantly produces single-label predictions. In contrast, \citet{yung-etal-2022-label} and \citet{costa2024exploring} also use the multi-label DiscoGeM corpus but convert its annotations into a single-label format during training. Thus, to date, no previous work has truly captured the full multi-label representation of implicit discourse relations - with the exception of our work on multi-lingual IDRR~\citep{costa2025multi-lingual}, where we consider multi-label IDRR in a multi-lingual setting incorporating hierarchical learning in the training of our classification model and comparing it against LLMs via direct prompting with few-shot learning. Similar to other works~\citep{chan-etal-2024-exploring,yung-etal-2024-prompting}, we found that prompting LLMs leads to worst results compared to the fine-tuning of PLMs such as RoBERTa in the context of IDRR. Therefore, we do not consider prompting LLMs in this paper.


\section{Our Approach}
\label{sec:approach}

In this work, we introduce a novel multi-task classification model for IDRR that simultaneously predicts sense distributions across all three levels in the PDTB~3.0 framework. As illustrated in Figure~\ref{fig:model}, the model processes a concatenated pair of discourse arguments (ARG1+ARG2) using RoBERTa~\citep{liu2019roberta} as the PLM encoder\footnote{We also experimented with other PLMs, but RoBERTa achieved the best performance (see Section~\ref{sec:model-selection}).}. The resulting contextualized embedding is passed through a linear transformation and dropout layer and subsequently fed to three distinct classification heads - each corresponding to a sense level in the hierarchy. In the multi-label setting, each head outputs a probability distribution over the available senses at its respective level. In the single-label setting, we apply an additional pooling layer that selects the sense with the highest probability from each distribution. We use the Adam optimizer~\citep{diederik2015adam} to minimize the loss function, which we calculate as the sum of the individual losses of each classification head.

\begin{figure}
    \centering
    \begin{tikzpicture}
        \tikzset{every node/.style={font=\small}}
    
        \draw[fill=new_green!40, rounded corners=0.1cm] (1,7) rectangle (3,7.5) node[pos=0.5] {ARG1};
    
        \draw[fill=new_green!40, rounded corners=0.1cm] (4.5,7) rectangle (6.5,7.5) node[pos=0.5] {ARG2};
    
        \draw[->] (3,7.25) -- (3.55,7.25);
    
        \draw[->] (4.5,7.25) -- (3.95,7.25);
    
        \draw (3.75,7.25) circle (0.2);
        \node[font=\footnotesize] at (3.75,7.25) {+};
    
        \draw[->] (3.75,7.05) -- (3.75,5.5);
    
        \draw[fill=new_yellow!40, rounded corners=0.1cm] (0,5.5) rectangle (7.5,6.5) node[pos=0.5] {RoBERTa-base};
    
        \draw[->] (3.75,5.5) -- (3.75,5);
    
        \draw[fill=new_yellow!40, rounded corners=0.1cm] (0,4.5) rectangle (7.5,5) node[pos=0.5] {Linear + Dropout Layer};

        \draw[color=primaryLight, rounded corners=0.1cm] (0,0.9) rectangle (2.2,3.6);
        \node[anchor=west, align=left, color=primaryLight] at (0,4.05) {Class Head \\ Level-1};

        \draw[->] (1.85,4.5) -- (1.85,3.6);

        \draw[fill=new_yellow!40, rounded corners=0.1cm] (0.1,3) rectangle (2.1,3.5) node[pos=0.5] {Linear Layer};

        \draw[->] (1.1,3) -- (1.1,2.5);

        \draw[fill=new_yellow!40, rounded corners=0.1cm] (0.1,2) rectangle (2.1,2.5) node[pos=0.5] {Norm. Layer};

        \draw[->] (1.1,2) -- (1.1,1.5);

        \draw[dashed, fill=new_yellow!40, rounded corners=0.1cm] (0.1,1.5) rectangle (2.1,1) node[pos=0.5] {Pool. Layer};

        \draw[->] (1.1,0.9) -- (1.1,0.5);

        \draw[fill=new_red!40, rounded corners=0.1cm] (0,0) rectangle (2.2,0.5) node[pos=0.5] {L1 Sense(s)};

        \draw[color=primaryLight, rounded corners=0.1cm] (2.65,0.9) rectangle (4.85,3.6);
        \node[anchor=west, align=left, color=primaryLight] at (2.65,4.05) {Class Head \\ Level-2};

        \draw[->] (4.5,4.5) -- (4.5,3.6);

        \draw[fill=new_yellow!40, rounded corners=0.1cm] (2.75,3) rectangle (4.75,3.5) node[pos=0.5] {Linear Layer};

        \draw[->] (3.75,3) -- (3.75,2.5);

        \draw[fill=new_yellow!40, rounded corners=0.1cm] (2.75,2) rectangle (4.75,2.5) node[pos=0.5] {Norm. Layer};

        \draw[->] (3.75,2) -- (3.75,1.5);

        \draw[dashed, fill=new_yellow!40, rounded corners=0.1cm] (2.75,1.5) rectangle (4.75,1) node[pos=0.5] {Pool. Layer};

        \draw[->] (3.75,0.9) -- (3.75,0.5);

        \draw[fill=new_red!40, rounded corners=0.1cm] (2.65,0) rectangle (4.85,0.5) node[pos=0.5] {L2 Sense(s)};

        \draw[color=primaryLight, rounded corners=0.1cm] (5.3,0.9) rectangle (7.5,3.6);
        \node[anchor=west, align=left, color=primaryLight] at (5.3,4.05) {Class Head \\ Level-3};

        \draw[->] (7.15,4.5) -- (7.15,3.6);

        \draw[fill=new_yellow!40, rounded corners=0.1cm] (5.4,3) rectangle (7.4,3.5) node[pos=0.5] {Linear Layer};

        \draw[->] (6.4,3) -- (6.4,2.5);

        \draw[fill=new_yellow!40, rounded corners=0.1cm] (5.4,2) rectangle (7.4,2.5) node[pos=0.5] {Norm. Layer};

        \draw[->] (6.4,2) -- (6.4,1.5);

        \draw[dashed, fill=new_yellow!40, rounded corners=0.1cm] (5.4,1.5) rectangle (7.4,1) node[pos=0.5] {Pool. Layer};

        \draw[->] (6.4,0.9) -- (6.4,0.5);

        \draw[fill=new_red!40, rounded corners=0.1cm] (5.3,0) rectangle (7.5,0.5) node[pos=0.5] {L3 Sense(s)};
    \end{tikzpicture}
    \caption{Architecture of our multi-task classification model for IDRR. Given a pair of discourse arguments as input, the model generates one output per sense level in the PDTB~3.0 framework. In the multi-label setting, each output is a probability distribution over the senses at the corresponding level. In the single-label setting, the sense with the highest probability is selected via the pooling layers (indicated by dashed lines).}
    \label{fig:model}
\end{figure}

\paragraph{Multi-Label Classification.} For each multi-label classification head, we compute the loss using the mean absolute error (MAE) loss function (see Equation~\ref{eq:multi-loss} in Appendix~\ref{apx:loss}), which we found to yield better performance (as detailed in Section~\ref{sec:model-selection}). Following previous work in multi-label classification for NLP~\citep{pyatkin2023design,van2024annotator}, we evaluate model performance in this setting using the Jensen-Shannon (JS) distance~\citep{lin1991divergence} to measure the similarity between the predicted and target probability distributions.

\paragraph{Single-Label Classification.} In the single-label setting, we compute the loss for each classification head using the cross-entropy (CE) loss function (see Equation~\ref{eq:single-loss} in Appendix~\ref{apx:loss}), which emphasizes the correct classification of the highest probability sense label (as discussed in Section~\ref{sec:model-selection}). We evaluate model performance in this setting using the weighted F1-score based on the majority label in the predicted and target sense distributions.


\section{Data Preparation}

We used two different discourse annotated corpora in this work: DiscoGeM~\citep{scholman-etal-2022-discogem} and PDTB~3.0~\citep{corpus-pdtb-3}. Our model was trained exclusively on the DiscoGeM corpus and evaluated on both DiscoGeM (for multi-label and single-label classification) and on PDTB~3.0 (for single-label classification). Although the two corpora differ in annotation methodologies, both follow the same discourse framework~\citep{webber2019penn}. This shared foundation allowed us to directly evaluate the transfer learning potential of training our model on DiscoGeM and evaluating it on PDTB~3.0 in single-label IDDR.


\subsection{DiscoGeM}
\label{sec:disco-data}

The DiscoGeM corpus contains 6,807 implicit discourse relations drawn from four textual genres: 2,800 relations from political texts, 3,060 from literary texts, 645 from encyclopedic texts and 302 relations extracted from the PDTB~3.0 corpus. Each relation was independently annotated by at least 10 crowdworkers and the resulting annotations were aggregated to form a probability distribution over 29 discourse senses in the PDTB~3.0 framework - the \textsc{Belief} and the \textsc{Speech-Act} level-2 senses were not annotated in DiscoGeM.

Since the DiscoGeM corpus did not include the \textsc{Cause+Belief} sense from the standard set of 14 level-2 senses proposed by \citet{kim-etal-2020-implicit} for IDRR classification in the PDTB~3.0 framework, we replaced it with the \textsc{Similarity} sense (the most frequent among the remaining level-2 senses) to preserve a 14-label set. In addition to classifying senses at level-1 and level-2, we also incorporated level-3 sense labels when available. When no level-3 sense was available, we defaulted to their corresponding level-2 sense as replacement. Table~\ref{tab:multi-statistics} in Appendix~\ref{apx:statistics} shows the distribution of senses across all levels in the original DiscoGeM annotations and in our adapted label set.

For single-label classification, we replaced the multi-label sense distribution of each discourse relation by the sense with the highest score - referred to as the majority label. To reduce computational complexity and maintain consistency across model evaluations, we used the same training, validation and testing splits in both the multi-label and the single-label settings. We split 70\% of the corpus for training, 10\% for validation and 20\% for testing. To ensure a balanced distribution, we kept the same distribution of majority labels across all data splits (see Figure~\ref{fig:discogem-label-distribution} in Appendix~\ref{apx:statistics}). We opted for a fixed split of the data, instead of multiple folds for cross-validation, to streamline the experimentation process as we conducted an extensive number of experiments.


\subsection{PDTB~3.0}
\label{sec:pdtb-data}

The PDTB~3.0 corpus was annotated by expert annotators and consists of 53,631 discourse relations extracted from Wall Street Journal news articles - from which 21,827 are implicit. Since there are currently no other benchmarks in multi-label IDRR using DiscoGeM and the majority of research in IDRR relies on the single-label annotated PDTB~3.0 corpus, we also evaluated our model on the traditional single-label classification task using different test splits of the PDTB~3.0 corpus through transfer learning.

To allow transfer learning between the two corpora, we used the same set of 14 level-2 senses described in Section~\ref{sec:disco-data}. Following the common approach to IDRR, we only kept the level-1 and level-2 senses in the PDTB~3.0. To compare our work against SOTA models in single-label IDRR, we replicated the two commonly used Lin~\citep{lin-etal-2009-recognizing} and Ji~\citep{ji-eisenstein-2015-one} test splits - the former uses section 23 in the PDTB~3.0 corpus as the test set, while the latter uses sections 21-22. However, since recent works highlighted the limitations of using such small test sets to draw meaningful conclusions~\citep{shi-demberg-2017-need,kim-etal-2020-implicit}, we also generated test splits following the cross-validation scheme proposed by \citet{kim-etal-2020-implicit}. Table~\ref{tab:single-statistics-test} in Appendix~\ref{apx:statistics} shows the total number of level-2 sense instances on the different single-label test sets.


\subsection{Corpora Agreement}
\label{sec:agreement}

Since a total of 302 implicit discourse relations taken from the PDTB~3.0 were also independently annotated in DiscoGeM, we calculated the annotation agreement between the two corpora over this overlapping set of relations. Table~\ref{tab:agreement} reports the number of relations where the PDTB~3.0 reference label matches the majority label (top-1) in DiscoGeM, or appears within its top-3, top-5, or top-10 majority labels. The fact that for 28.8\% of the jointly annotated relations, the reference label from the PDTB 3.0 corpus was not selected by at least one of the annotators of DiscoGeM illustrates the variability in human interpretation of implicit discourse relations - even when annotations are based on the same underlying framework.

\begin{table}
    \centering
    \renewcommand{\arraystretch}{1.2}
    \resizebox{\columnwidth}{!}{
        \begin{tabular}{|c|c|c|c|}
            \hline
            \hline
            Top-1 & Top-3 & Top-5 & Top-10 \\
            \hline
            $92 \; (30.5\%)$ & $164 \; (54.3\%)$ & $197 \; (65.2\%)$ & $215 \; (71.2\%)$ \\
            \hline
            \hline
        \end{tabular}
    }
    \caption{Number of instances where the reference label in the PDTB~3.0 was found within the top-k labels in DiscoGeM for the set of 302 co-annotated relations.}
    \label{tab:agreement}
\end{table}


\section{Experiments}
\label{sec:experiments}

To optimize our model (illustrated in Figure~\ref{fig:model}), we explored multiple model configurations and loss functions using the validation set of DiscoGeM. All of the results reported in the following sections are the average scores of three different runs with different random starts for 10 epochs and with a batch size of 16. We ran our experiments on a 32-core compute node with 512GB of RAM. All of the code used can be found on GitHub\footnote{\url{https://github.com/nelsonfilipecosta/Implicit-Discourse-Relation-Recognition}}.

\begin{table*}
    \centering
    \renewcommand{\arraystretch}{1.2}
    \resizebox{\textwidth}{!}{
        \begin{tabular}{|c|c|ccc|ccc|}
            \hline
            \hline
            \multirow{2}{*}{PLM} & \multirow{2}{*}{Loss} & \multicolumn{3}{c|}{JS Distance \scalebox{0.9}{$\searrow$} (Multi-Label)} & \multicolumn{3}{c|}{F1-Score \scalebox{0.9}{$\nearrow$} (Single-Label)} \\
            \cline{3-8}
             &  & \multicolumn{1}{c}{Level-1} & \multicolumn{1}{c}{Level-2} & \multicolumn{1}{c|}{Level-3} & \multicolumn{1}{c}{Level-1} & \multicolumn{1}{c}{Level-2} & \multicolumn{1}{c|}{Level-3} \\
            \hline
            \hline
            \multirow{2}{*}{BERT} & MAE & $0.314 \pm 0.004$ & $0.461 \pm 0.007$ & $0.535 \pm 0.004$ & $63.34 \pm 1.41$ & $46.44 \pm 0.82$ & $39.20 \pm 1.82$ \\
             & CE & $0.328\pm 0.004$ & $0.563 \pm 0.002$ & $0.634 \pm 0.004$ & $63.49 \pm 1.38$ & $51.42 \pm 0.58$ & $47.65 \pm 0.92$ \\
            \hline
            \multirow{2}{*}{DistilBERT} & MAE & $0.330 \pm 0.006$ & $0.472 \pm 0.003$ & $0.543 \pm 0.003$ & $61.29 \pm 0.96$ & $39.29 \pm 0.66$ & $31.75 \pm 0.55$ \\
             & CE & $0.341 \pm 0.005$ & $0.571 \pm 0.006$ & $0.639 \pm 0.003$ & $60.28 \pm 0.58$ & $45.55 \pm 1.52$ & $41.19 \pm 1.50$ \\
            \hline
            \multirow{2}{*}{RoBERTa} & MAE & \bm{$0.304 \pm 0.010$} & \bm{$0.448 \pm 0.006$} & \bm{$0.530 \pm 0.004$} & $65.12 \pm 1.20$ & $51.81 \pm 1.54$ & $42.99 \pm 1.09$ \\
             & CE & $0.317 \pm 0.003$ & $0.561 \pm 0.003$ & $0.630 \pm 0.005$ & \bm{$65.48 \pm 0.82$} & \bm{$53.51 \pm 0.39$} & \bm{$49.55 \pm 0.11$} \\
            \hline
            \multirow{2}{*}{DistilRoBERTa} & MAE & $0.318 \pm 0.001$ & $0.463 \pm 0.003$ & $0.539 \pm 0.004$ & $61.30 \pm 0.38$ & $42.75 \pm 1.63$ & $34.96 \pm 1.29$ \\
             & CE & $0.329 \pm 0.002$ & $0.564 \pm 0.005$ & $0.636 \pm 0.003$ & $65.02 \pm 0.69$ & $52.83 \pm 0.71$ & $47.94 \pm 0.88$ \\
            \hline
            \hline
        \end{tabular}
    }
    \caption{Results of experimenting with different pre-trained language models (PLMs) and different loss functions in multi-label classification (JS distance) and in single-label classification (weighted F1-score). The results were averaged across three different runs with random starts. Values in bold show the best score for each metric.}
    \label{tab:model-selection-1}
\end{table*}


\subsection{Model Selection}
\label{sec:model-selection}

Motivated by prior work suggesting that PLMs trained with a next sentence prediction (NSP) objective perform better on IDRR~\citep{shi-demberg-2019-next}, we compared BERT~\citep{devlin-etal-2019-bert}, which includes the NSP objective, against RoBERTa~\citep{liu2019roberta}, which omits it. RoBERTa has also been widely adopted in SOTA single-label IDRR models~\citep{long-webber-2022-facilitating,liu-strube-2023-annotation,zhao-etal-2023-infusing,zeng-etal-2024-global,long2024leveraging}, making it a strong baseline. To further explore efficiency-performance trade-offs, we also evaluated distilled variants of both models~\citep{sanh2019distilbert}. Each PLM configuration was evaluated using both the mean absolute error (MAE) and cross-entropy (CE) loss functions. The results of these experiments are summarized in Table~\ref{tab:model-selection-1}.

As shown in Table~\ref{tab:model-selection-1}, RoBERTa consistently outperforms BERT across both loss functions, yielding lower Jensen-Shannon (JS) distances in the multi-label setting and higher F1-scores in the single-label setting across all sense levels. These results suggest that the performance gains of using larger models outweigh the previously reported advantages of using models pre-trained on the NSP objective in IDRR~\citep{shi-demberg-2019-next}. In addition, both BERT and RoBERTa outperform their respective distilled variants, reaffirming the benefit of using full-sized models for this task. The results in Table~\ref{tab:model-selection-1} also show that the choice of loss function plays an important role. Models trained with MAE consistently achieve lower JS distances in the multi-label setting, while model trained with CE achieve higher F1-scores in the single-label setting. We present a more detailed analysis on the performance of each loss function in Appendix~\ref{apx:loss}, as well as the results of training the different models with the mean squared error (MSE) and the Huber loss - which led to slightly worse results as shown in Table~\ref{tab:model-selection-2} of the appendix. Based on these findings, we selected RoBERTa as the backbone encoder of our model, using MAE loss for multi-label classification and CE loss for single-label classification.


\subsection{Fine-Tuning}

\begin{table*}
    \centering
    \renewcommand{\arraystretch}{1.2}
    \resizebox{\textwidth}{!}{
        \begin{tabular}{|c|c|ccc|ccc|}
            \hline
            \hline
            \multirow{2}{*}{LR} & \multirow{2}{*}{Decay} & \multicolumn{3}{c|}{JS Distance \scalebox{0.9}{$\searrow$} (Multi-Label with MAE)} & \multicolumn{3}{c|}{F1-Score \scalebox{0.9}{$\nearrow$} (Single-Label with CE)} \\
            \cline{3-8}
             &  & \multicolumn{1}{c}{Level-1} & \multicolumn{1}{c}{Level-2} & \multicolumn{1}{c|}{Level-3} & \multicolumn{1}{c}{Level-1} & \multicolumn{1}{c}{Level-2} & \multicolumn{1}{c|}{Level-3} \\
            \hline
            \hline
            $1e^{-4}$ & $-$ & $0.386 \pm 0.004$ & $0.509 \pm 0.003$ & $0.569 \pm 0.005$ & $36.07 \pm 1.67$ & $21.61 \pm 0.58$ & $10.05 \pm 0.99$ \\
            $5e^{-5}$ & $-$ & $0.339 \pm 0.041$ & $0.478 \pm 0.039$ & $0.554 \pm 0.016$ & $64.71 \pm 2.39$ & $49.76 \pm 1.65$ & $45.52 \pm 1.59$ \\
            \hdashline
            \multirow{3}{*}{$1e^{-5}$} & $-$ & $0.304 \pm 0.010$ & $0.448 \pm 0.006$ & $0.530 \pm 0.004$ & \bm{$65.48 \pm 0.82$} & $53.51 \pm 0.39$ & $49.55 \pm 0.11$ \\
             & Linear & $0.305 \pm 0.009$ & $0.459 \pm 0.007$ & $0.547 \pm 0.004$ & $64.70 \pm 0.35$ & $54.32 \pm 2.15$ & $48.96 \pm 0.81$ \\
             & CosAn & \bm{$0.299 \pm 0.004$} & \bm{$0.447 \pm 0.008$} & \bm{$0.529 \pm 0.008$} & $65.15 \pm 0.61$ & $53.99 \pm 1.17$ & $49.56 \pm 1.25$ \\ 
            \hdashline
            \multirow{3}{*}{$5e^{-6}$} & $-$ & $0.303 \pm 0.006$ & $0.461 \pm 0.005$ & $0.548 \pm 0.005$ & $65.45 \pm 0.36$ & $53.67 \pm 0.45$ & $49.77 \pm 1.15$ \\
             & Linear & $0.315 \pm 0.006$ & $0.480 \pm 0.007$ & $0.566 \pm 0.007$ & $64.84 \pm 1.14$ & \bm{$55.86 \pm 1.10$} & \bm{$50.34 \pm 1.55$} \\
             & CosAn & $0.306 \pm 0.003$ & $0.469 \pm 0.002$ & $0.558 \pm 0.004$ & $65.40 \pm 1.04$ & $54.97 \pm 1.85$ & $49.62 \pm 1.26$ \\
            \hdashline
            $1e^{-6}$ & $-$ & $0.348 \pm 0.003$ & $0.522 \pm 0.005$ & $0.613 \pm 0.003$ & $60.66 \pm 1.38$ & $47.66 \pm 0.64$ & $40.96 \pm 0.48$ \\
            \hline
            \hline
        \end{tabular}
    }
    \caption{Results of experimenting with different learning rates and decay functions in multi-label classification (JS distance) using the MAE loss and in single-label classification (weighted F1-score) using the CE loss. The results were averaged across three different runs with random starts. Values in bold show the best score for each metric.}
    \label{tab:fine-tuning-mixed}
\end{table*}

We fine-tuned our model separately with different learning rates for multi-label and single-label IDRR. Table~\ref{tab:fine-tuning-mixed} shows the results of fine-tuning our model with the MAE loss in the multi-label setting and with the CE loss in the single-label setting. Under both settings, we observed performance improvements with learning rates of $1 \times 10^{-5}$ and $5 \times 10^{-6}$. To further enhance model performance with these two learning rates, we also experimented incorporating two decay functions: linear decay and cosine annealing with warm restarts~\citep{ilya2017sgdr}. For linear decay, the learning rate was gradually reduced over the first 5 epochs to half its initial value. In the cosine annealing schedule, the learning rate oscillated between its original value and half that value over two complete cycles across 10 epochs. As shown in Table~\ref{tab:fine-tuning-mixed}, the cosine annealing strategy with a learning rate of $1 \times 10^{-5}$ achieved the best results for multi-label classification, while the linear decay schedule with a learning rate of $5 \times 10^{-6}$ proved most effective in the single-label setting.


\section{Results}
\label{sec:results}

In this section, we present the results of our experiments in two parts. Section~\ref{sec:results_disco} reports the performance of our models on the DiscoGeM test set for both multi-label and single-label IDRR and Section~\ref{sec:results_pdtb} presents the transfer learning results of our single-label model evaluated on various test splits of the PDTB~3.0 corpus.


\subsection{Results on DiscoGeM}
\label{sec:results_disco}

\begin{table*}
    \centering
    \renewcommand{\arraystretch}{1.2}
    \resizebox{\textwidth}{!}{
        \begin{tabular}{|c|ccc|ccc|}
            \hline
            \hline
            \multirow{2}{*}{Models} & \multicolumn{3}{c|}{JS Distance \scalebox{0.9}{$\searrow$} (Multi-Label)} & \multicolumn{3}{c|}{F1-Score \scalebox{0.9}{$\nearrow$} (Single-Label)} \\
            \cline{2-7}
            & \multicolumn{1}{c}{Level-1} & \multicolumn{1}{c}{Level-2} & \multicolumn{1}{c|}{Level-3} & \multicolumn{1}{c}{Level-1} & \multicolumn{1}{c}{Level-2} & \multicolumn{1}{c|}{Level-3} \\
            \hline
            \hline
            Baseline & $0.519 \pm 0.002$ & $0.636 \pm 0.001$ & $0.714 \pm 0.002$ & $34.72 \pm 1.63$ & $22.42 \pm 1.34$ & $18.81 \pm 0.80$ \\
            \hdashline
           ~\cite{yung-etal-2022-label} & $-$ & $-$ & $-$ & $-$ & $23.66 \pm 1.19$ & $-$ \\
           ~\cite{costa2024exploring} & $-$ & $-$ & $-$ & $-$ & $-$ & \bm{$51.38 \pm n/a$} \\
            \hdashline
            Ours w/ MAE & \bm{$0.299 \pm 0.002$} & \bm{$0.446 \pm 0.003$} & \bm{$0.523 \pm 0.002$} & $65.39 \pm 0.54$ & $50.13 \pm 1.54$ & $41.60 \pm 0.59$ \\
            Ours w/ CE & $0.323 \pm 0.003$ & $0.564 \pm 0.003$ & $0.634 \pm 0.002$ & \bm{$65.89 \pm 1.35$} & \bm{$55.99 \pm 1.73$} & \bm{$50.82 \pm 1.26$} \\
            \hline
            \hline
        \end{tabular}
    }
    \caption{Final results on the test set of DiscoGeM in multi-label classification (JS distance) and in single-label classification (weighted F1-score). The results were averaged across three different runs with random starts. Values in bold show the best score for each metric.}
    \label{tab:results-discogem}
\end{table*}

Table~\ref{tab:results-discogem} presents the performance of our models (trained with MAE and CE loss functions) on the DiscoGeM test set for both multi-label and single-label IDRR. Results in the multi-label setting are reported using Jensen-Shannon (JS) distance, while results in the single-label setting are reported using weighted F1-score. As there is currently no prior work on full multi-label IDRR, we include a baseline for comparison. This baseline emulates the DiscoGeM annotation protocol by generating ten single-label predictions per relation, sampled from the probability distribution of each sense at each level, and then averaging them into a probability distribution. In the single-label setting, we compare our results against prior work evaluated on DiscoGeM~\citep{yung-etal-2022-label,costa2024exploring}. As shown in Table~\ref{tab:results-discogem}, our model consistently outperforms the random baseline across all levels and achieves a substantial improvement over the current SOTA in level-2 classification~\citep{yung-etal-2022-label}. At level-3, our model performs at par with the best reported results from \citet{costa2024exploring}. It is worth noting that their model was specifically trained to predict level-3 senses, whereas our model is designed to generalize across all three sense levels simultaneously.


\subsection{Transfer Learning Results on PDTB~3.0}
\label{sec:results_pdtb}

The results of transfer learning from DiscoGeM to PDTB~3.0 on single-label IDRR are presented in Table~\ref{tab:results-pdtb}. We compare our model against SOTA approaches in this task~\citep{long-webber-2022-facilitating,liu-strube-2023-annotation,zhao-etal-2023-infusing,zeng-etal-2024-global,long2024leveraging} and \citet{yung-etal-2022-label}. As shown in Table~\ref{tab:results-pdtb}, our model has a lower performance than those trained on PDTB~3.0. However, it is important to emphasize that our model was trained exclusively on DiscoGeM and did not see any PDTB~3.0 data in its training. The lower performance in this zero-shot transfer learning setting, where no fine-tuning is done on the target corpus, can also be explained by thelimited agreement between the two corpora - as shown in Table~\ref{tab:agreement}, the majority sense in DiscoGeM only matched the PDTB~3.0 sense on $92$ $(30.5\%)$ of the co-annotated discourse relations. This also explains the contrast in performances shown in Tables~\ref{tab:results-discogem}~and~\ref{tab:results-pdtb} between our model and the one from \citet{yung-etal-2022-label}, which was trained on the two corpora. Our model performs significantly better when evaluated on DiscoGeM, but worst when evaluated on the PDTB~3.0. These findings suggest that, to improve cross-corpus generalization, future work should explore intermediate fine-tuning on PDTB~3.0 after pretraining on DiscoGeM.

\begin{table*}[!ht]
    \centering
    \renewcommand{\arraystretch}{1.2}
    \resizebox{\textwidth}{!}{
        \begin{tabular}{|c|cc|cc|cc|}
            \hline
            \hline
            \multirow{3}{*}{Models} & \multicolumn{6}{c|}{F1-Score (PDTB~3.0)} \\
            \cline{2-7}
             & \multicolumn{2}{c|}{Lin} & \multicolumn{2}{c|}{Ji} & \multicolumn{2}{c|}{Cross} \\
            \cline{2-7}
             & \multicolumn{1}{c}{Level-1} & \multicolumn{1}{c|}{Level-2} & \multicolumn{1}{c}{Level-1} & \multicolumn{1}{c|}{Level-2} & \multicolumn{1}{c}{Level-1} & \multicolumn{1}{c|}{Level-2} \\
            \hline
            \hline
            \citet{yung-etal-2022-label} & $-$ & $-$ & $-$ & $38.07 \pm 2.25$ & $-$ & $-$ \\
            \citet{long-webber-2022-facilitating} & $-$ & $-$ & $70.05 \pm n/a$ & \bm{$57.62 \pm n/a$} & $-$ & $-$ \\
            \citet{liu-strube-2023-annotation} & $-$ & $-$ & $71.15 \pm 0.47$ & $54.92 \pm 0.81$ & \bm{$70.06 \pm 1.72$} & \bm{$55.26 \pm 1.32$} \\
            \citet{zhao-etal-2023-infusing} & $-$ & $-$ & $69.06 \pm n/a$ & $52.73 \pm n/a$ & $-$ & $-$ \\
            \citet{zeng-etal-2024-global} & $-$ & $-$ & \bm{$71.59 \pm n/a$} & $56.50 \pm n/a$ & $-$ & $-$ \\
            \citet{long2024leveraging} & $-$ & $-$ & $71.19 \pm n/a$ & $52.91 \pm n/a$ & $-$ & $-$ \\
            \hdashline
            Ours w/ MAE & $47.17 \pm 1.16$ & $27.89 \pm 1.21$ & $43.88 \pm 1.06$ & $26.29 \pm 1.71$ & $45.53 \pm 1.23$ & $29.30 \pm 1.49$ \\
            Ours w/ CE & \bm{$50.44 \pm 0.81$} & \bm{$34.38 \pm 0.86$} & $49.43 \pm 1.47$ & $33.11 \pm 1.57$ & $49.72 \pm 1.45$ & $33.21 \pm 1.67$ \\
            \hline
            \hline
        \end{tabular}
    }
    \caption{Transfer learning results on different test splits of the PDTB~3.0 in single-label classification (weighted F1-score). The Lin and Ji results were averaged across three different runs with random starts, while the Cross results were averaged across all 12 folds. Values in bold show the best score for each level.}
    \label{tab:results-pdtb}
\end{table*}


\section{Analysis}

To further evaluate the performance of our model, we conduct two additional analyses. In Section~\ref{sec:individual}, we perform a per-sense evaluation to assess the ability of the model to correctly classify individual senses at both level-1 and level-2. In Section~\ref{sec:coherence}, we evaluate the cross-level coherence of the predictions by examining how well the predicted labels at level-1 and level-2 align with the sense hierarchy defined in the PDTB~3.0 framework. For ease of interpretation, we consider only single-label results.


\subsection{Per-Sense Results}
\label{sec:individual}

To identify disparities in performance between individual labels and to assess how well the model is able to generalize across senses, we evaluate the single-label predictions of our model at level-1 and level-2. Table~\ref{tab:individual_results} shows the weighted F1-score of each predicted sense in the test set of DiscoGeM and in the different test sets of the PDTB~3.0.

\begin{table*}[!ht]
    \centering
    \renewcommand{\arraystretch}{1.2}
    \resizebox{\textwidth}{!}{
        \begin{tabular}{|c|c|c|c|c|c|c|c|c|c|}
            \hline
            \hline
            \multirow{2}{*}{Level-1} & DiscoGeM & \multicolumn{3}{c|}{PDTB~3.0} & \multirow{2}{*}{Level-2} & DiscoGeM & \multicolumn{3}{c|}{PDTB~3.0} \\
            \cline{2-5}
            \cline{7-10}
             & \multicolumn{1}{c|}{Test} & \multicolumn{1}{c|}{Lin} & \multicolumn{1}{c|}{Ji} & \multicolumn{1}{c|}{Cross} &  & \multicolumn{1}{c|}{Test} & \multicolumn{1}{c|}{Lin} & \multicolumn{1}{c|}{Ji} & \multicolumn{1}{c|}{Cross} \\
            \hline
            \hline
            \multirow{2}{*}{\textsc{Temporal}} & \multirow{2}{*}{$61.27 \pm 1.34$} & \multirow{2}{*}{$4.32 \pm 1.58$} & \multirow{2}{*}{$15.75 \pm 4.50$} & \multirow{2}{*}{$16.23 \pm 4.86$} & \textsc{Synchronous} & $0.00 \pm 0.00$ & $0.00 \pm 0.00$ & $0.00 \pm 0.00$ & $0.00 \pm 0.00$ \\
             & & & & & \textsc{Asynchronous} & $57.87 \pm 0.84$ & $6.01 \pm 2.05$ & $22.02 \pm 4.42$ & $23.16 \pm 6.65$ \\
            \hline
            \multirow{3}{*}{\textsc{Contingency}} & \multirow{3}{*}{$58.67 \pm 1.83$} & \multirow{3}{*}{$36.12 \pm 7.17$} & \multirow{3}{*}{$34.27 \pm 8.24$} & \multirow{3}{*}{$41.08 \pm 2.87$} & \textsc{Cause} & $65.36 \pm 0.12$ & $47.21 \pm 1.93$ & $46.40 \pm 2.47$ & $48.93 \pm 2.07$ \\
             & & & & & \textsc{Condition} & $-$ & $0.00 \pm 0.00$ & $0.00 \pm 0.00$ & $0.00 \pm 0.00$ \\
             & & & & & \textsc{Purpose} & $-$ & $0.00 \pm 0.00$ & $0.00 \pm 0.00$ & $0.00 \pm 0.00$ \\
            \hline
            \multirow{3}{*}{\textsc{Comparison}} & \multirow{3}{*}{$39.31 \pm 3.15$} & \multirow{3}{*}{$32.65 \pm 1.69$} & \multirow{3}{*}{$37.67 \pm 5.20$} & \multirow{3}{*}{$33.72 \pm 2.34$} & \textsc{Concession} & $36.24 \pm 4.55$ & $29.70 \pm 5.09$ & $27.92 \pm 1.75$ & $19.46 \pm 3.70$ \\
             & & & & & \textsc{Contrast} & $4.13 \pm 2.94$ & $11.10 \pm 7.92$ & $12.70 \pm 9.35$ & $10.43 \pm 6.61$ \\
             & & & & & \textsc{Similarity} & $0.00 \pm 0.00$ & $0.00 \pm 0.00$ & $0.00 \pm 0.00$ & $0.00 \pm 0.00$ \\
            \hline
            \multirow{6}{*}{\textsc{Expansion}} & \multirow{6}{*}{$75.39 \pm 1.15$} & \multirow{6}{*}{$64.44 \pm 0.66$} & \multirow{6}{*}{$62.69 \pm 0.38$} & \multirow{6}{*}{$63.48 \pm 1.16$} & \textsc{Conjunction} & $52.56 \pm 0.66$ & $42.30 \pm 1.26$ & $39.91 \pm 1.97$ & $44.45 \pm 3.04$ \\
             & & & & & \textsc{Equivalence} & $-$ & $0.00 \pm 0.00$ & $0.00 \pm 0.00$ & $0.00 \pm 0.00$ \\
             & & & & & \textsc{Instantiation} & $44.84 \pm 0.18$ & $12.15 \pm 9.12$ & $17.90 \pm 12.72$ & $15.62 \pm 10.49$ \\
             & & & & & \textsc{Level-of-Detail} & $46.85 \pm 3.28$ & $33.61 \pm 2.90$ & $29.32 \pm 2.70$ & $29.27 \pm 3.06$ \\
             & & & & & \textsc{Manner} & $-$ & $0.00 \pm 0.00$ & $0.00 \pm 0.00$ & $0.00 \pm 0.00$ \\
             & & & & & \textsc{Substitution} & $0.00 \pm 0.00$ & $0.00 \pm 0.00$ & $0.00 \pm 0.00$ & $0.00 \pm 0.00$ \\
            \hline
            \hline
        \end{tabular}
    }
    \caption{Individual per-sense results for each sense in level-1 and level-2 on the test set of DiscoGeM and on different test splits of the PDTB~3.0 in single-label classification (weighted F1-score). Senses marked with "$-$" were not present in the DiscoGeM test set. The DiscoGeM and the PDTB~3.0 Lin and Ji results were averaged across three different runs with random starts, while the PDTB~3.0 Cross results were averaged across all 12 folds.}
    \label{tab:individual_results}
\end{table*}

As shown in Table~\ref{tab:individual_results}, the per-sense performance of the model on the DiscoGeM test set at both level-1 and level-2 generally reflects the distribution of senses in the corpus (see Table~\ref{tab:multi-statistics} in Appendix~\ref{apx:statistics}). Note that some certain senses — \textsc{Condition}, \textsc{Purpose}, \textsc{Equivalence} and \textsc{Manner} — never appear as the majority label of relation in DiscoGeM and are therefore absent from the test set (see Figure~\ref{fig:discogem-label-distribution} and Table~\ref{tab:single-statistics-test} in Appendix~\ref{apx:statistics}). Unsurprisingly, the model struggles to predict under-represented senses such as \textsc{Synchronous}, \textsc{Contrast}, \textsc{Similarity} and \textsc{Substitution}, which appear infrequently in the training data. Addressing this imbalance, potentially through targeted data augmentation, could improve generalization on these under-represented senses. The confusion matrices in Figures~\ref{fig:confusion_l1}~and~\ref{fig:confusion_l2} in Appendix~\ref{apx:confusion} provide further details into the per-sense performance on the test set of DiscoGeM. The performance of the model on the PDTB~3.0 test sets, also shown in Table~\ref{tab:individual_results}, is consistently lower than on DiscoGeM, mirroring the trends reported in Table~\ref{tab:results-discogem} and Table~\ref{tab:results-pdtb}. This performance gap is particularly pronounced for certain senses, such as \textsc{Temporal} at level\nobreakdash-1 and \textsc{Asynchronous} and \textsc{Instantiation} at level-2. Conversely, the model performs better at the level-2 \textsc{Contrast} sense. Since the model was not trained on any PDTB~3.0 data, these discrepancies can be explained by annotation inconsistencies across both corpora as highlighted in Section~\ref{sec:agreement}.


\subsection{Sense Coherence across Levels}
\label{sec:coherence}

\begin{table*}
    \centering
    \renewcommand{\arraystretch}{1.2}
    \resizebox{\textwidth}{!}{
        \begin{tabular}{|c|c|c|c|c|c|c|c|c|c|}
            \hline
            \hline
            \multirow{2}{*}{Level-1} & DiscoGeM & \multicolumn{3}{c|}{PDTB~3.0} & \multirow{2}{*}{Level-2} & DiscoGeM & \multicolumn{3}{c|}{PDTB~3.0} \\
            \cline{2-5}
            \cline{7-10}
             & \multicolumn{1}{c|}{Test (\%)} & \multicolumn{1}{c|}{Lin (\%)} & \multicolumn{1}{c|}{Ji (\%)} & \multicolumn{1}{c|}{Cross (\%)} &  & \multicolumn{1}{c|}{Test (\%)} & \multicolumn{1}{c|}{Lin (\%)} & \multicolumn{1}{c|}{Ji (\%)} & \multicolumn{1}{c|}{Cross (\%)} \\
            \hline
            \hline
            \multirow{2}{*}{\textsc{Temporal}} & \multirow{2}{*}{$92.94 \pm 2.93$} & \multirow{2}{*}{$93.27 \pm 6.31$} & \multirow{2}{*}{$90.93 \pm 4.51$} & \multirow{2}{*}{$92.48 \pm 4.93$} & \textsc{Synchronous} & $n/a$ & $n/a$ & $n/a$ & $n/a$ \\
             & & & & & \textsc{Asynchronous} & $90.00 \pm 1.44$ & $82.74 \pm 6.76$ & $82.44 \pm 11.34$ & $88.30 \pm 8.08$ \\
            \hline
            \multirow{3}{*}{\textsc{Contingency}} & \multirow{3}{*}{$98.91 \pm 0.20$} & \multirow{3}{*}{$100.00 \pm 0.00$} & \multirow{3}{*}{$99.74 \pm 0.45$} & \multirow{3}{*}{$100.00 \pm 0.00$} & \textsc{Cause} & $62.37 \pm 4.92$ & $38.19 \pm 6.45$ & $38.42 \pm 8.27$ & $29.93 \pm 1.32$ \\
             & & & & & \textsc{Condition} & $-$ & $n/a$ & $n/a$ & $n/a$ \\
             & & & & & \textsc{Purpose} & $-$ & $n/a$ & $n/a$ & $n/a$ \\
            \hline
            \multirow{3}{*}{\textsc{Comparison}} & \multirow{3}{*}{$86.31 \pm 1.22$} & \multirow{3}{*}{$73.81 \pm 13.21$} & \multirow{3}{*}{$68.09 \pm 10.47$} & \multirow{3}{*}{$58.45 \pm 4.12$} & \textsc{Concession} & $81.19 \pm 2.83$ & $89.15 \pm 8.10$ & $89.36 \pm 8.98$ & $97.93 \pm 1.52$ \\
             & & & & & \textsc{Contrast} & $91.67 \pm 14.43$ & $90.77\pm 10.09$ & $ 90.77 \pm 11.11$ & $92.68 \pm 8.19$ \\
             & & & & & \textsc{Similarity} & $n/a$ & $n/a$ & $n/a$ & $n/a$ \\
             
            \hline
            \multirow{6}{*}{\textsc{Expansion}} & \multirow{6}{*}{$70.70 \pm 1.26$} & \multirow{6}{*}{$37.91 \pm 10.82$} & \multirow{6}{*}{$65.02 \pm 12.30$} & \multirow{6}{*}{$53.01 \pm 2.21$} & \textsc{Conjunction} & $97.61 \pm 0.71$ & $97.14 \pm 0.03$ & $3.20 \pm 0.88$ & $96.69 \pm 0.96$ \\
            
             & & & & & \textsc{Equivalence} & $-$ & $n/a$ & $n/a$ & $n/a$ \\
             & & & & & \textsc{Instantiation} & $100.00 \pm 0.00$ & $100.00 \pm 0.00$ & $100.00 \pm 0.00$ & $100.00 \pm 0.00$ \\
             & & & & & \textsc{Level-of-Detail} & $99.82 \pm 0.31$ & $99.80 \pm 0.35$ & $100.00 \pm 0.00$ & $100.00 \pm 0.00$ \\
             & & & & & \textsc{Manner} & $-$ & $n/a$ & $n/a$ & $n/a$ \\
             & & & & & \textsc{Substitution} & $n/a$ & $n/a$ & $n/a$ & $n/a$ \\
            \hline
            \hline
        \end{tabular}
    }
    \caption{Percentage of the instances that were classified with coherent senses at level-1 and level-2 on the test set of DiscoGeM and on different test splits of the PDTB~3.0 in single-label classification. Senses marked with "$-$" were not present in the DiscoGeM test set and senses marked with "$n/a$" were never predicted by the model. The DiscoGeM and the PDTB~3.0 Lin and Ji results were averaged across three different runs with random starts, while the PDTB~3.0 Cross results were averaged across all 12 folds.}
    \label{tab:coherence}
\end{table*}

Table~\ref{tab:coherence} shows the percentage of times a sense at level-1 was predicted with a coherent sense at level-2 (and vice-versa) in the test set of DiscoGeM and in the different test sets of the PDTB~3.0. This enables us to examine the extent to which the predictions of the model are consistent with the hierarchical sense structure defined in the PDTB~3.0.

As Table~\ref{tab:coherence} shows, the \textsc{Contingency} sense at level-1 is very often predicted with a coherent level-2 sense (i.e., \textsc{Cause}, \textsc{Condition} or \textsc{Purpose}). Similarly, the level-2 senses \textsc{Instantiation}, \textsc{Level-of-Detail} and \textsc{Conjunction} are often predicted with a coherent level-1 sense. However, other senses, such as \textsc{Comparison} and \textsc{Expansion} at level-1 and \textsc{Cause} at level-2, are more often predicted with a contradicting sense. A possible explanation might be that these senses are inherently hard to distinguish and the model is biased towards the most represented sense. For instance, the level-2 senses \textsc{Cause} and \textsc{Asynchronous} are among the most often co-annotated pair of senses in the DiscoGeM and \textsc{Cause} appears approximately three times more often than \textsc{Asynchronous} (see Table~\ref{tab:multi-statistics} in Appendix~\ref{apx:statistics}), the model is likely more biased towards \textsc{Cause} when distinguishing between these two senses. One possible method to generate more coherent predictions across sense levels, would be to share the predictions of lower-level senses to help inform the prediction of higher-level senses within the model.


\section{Conclusion}

In this work, we proposed a novel multi-label framework for IDRR, addressing one of the most complex challenges in discourse analysis. We trained a multi-task classification model on the DiscoGeM corpus to simultaneously learn multi-label representations of discourse relations across all three sense levels in the PDTB~3.0 framework. Our model can also be adapted to the traditional single-label IDRR setting by selecting the sense with the highest probability in the multi-label representation. We conducted extensive experiments to identify optimal model configurations and loss functions in both settings. Our approach establishes a first benchmark on multi-label IDRR and achieves SOTA results in single-label IDRR using DiscoGeM. Finally, we evaluated our model on the PDTB~3.0 corpus in the single-label setting, presenting the first analysis of transfer learning between the DiscoGeM and PDTB~3.0 corpora for IDRR. Our results show that zero-shot direct transfer learning between both corpora still needs further research.

Future work should explore the application of data augmentation techniques, such as paraphrasing, to determine whether augmenting under-represented senses in the DiscoGeM could enhance the performance of the model on individual senses. Additionally, future research should investigate whether cascading information from higher-level to lower-level classification heads within the model could improve the coherence of its predictions across sense levels. Lastly, it would be worthwhile to examine whether a model trained to learn multi-label representations of discourse relations in DiscoGeM could be further fine-tuned on the PDTB~3.0 to achieve superior performance in single-label IDRR using the PDTB~3.0 corpus.


\section{Limitations}

Despite the promising results, there are a few limitations to our work. The first limitation of our work comes from the fact that the \textsc{CAUSE+BELIEF} sense was not annotated in the DiscoGeM corpus. Therefore, we could not use the standard 14-label set of second-level senses proposed for the PDTB 3.0 by \citet{kim-etal-2020-implicit} and widely used in literature. Instead, we replaced it by the \textsc{SIMILARITY} sense - the next most available sense in the DiscoGeM corpus. Ideally, we would have replaced the \textsc{Cause+Belief} sense with another sense under the \textsc{Contingency} level-1 sense, but they were all already considered.

Our choice of pre-trained language model to generate the embeddings for each pair of discourse arguments also imposes a limitation to our work. Due to computational and time limitations, we could not explore the fine-tuning of larger models, such as LLaMA~3 \citep{grattafiori2024llama}, nor explore prompting LLMs. However, in our other work \citep{costa2025multi-lingual}, we show that the use of LLMs via direct prompting with few-shot learning for IDRR does not lead to better results. Finally, we would like to acknowledge that the extensive experimental analysis conducted in this work would not have been possible without access to a high-performance computing facility - which entails a non-negligible carbon footprint which we did not monitor during any of our experiments.


\section*{Acknowledgements}

The authors would like to thank the anonymous reviewers for their comments. This work was financially supported by the Natural Sciences and Engineering Research Council of Canada (NSERC).


\bibliography{nelson}

\begin{thebibliography}{53}
\providecommand{\natexlab}[1]{#1}

\bibitem[{Asher and Lascarides(2003)}]{asher2003logics}
Nicholas Asher and Alex Lascarides. 2003.
\newblock \emph{{Logics of Conversation}}.
\newblock Cambridge University Press.

\bibitem[{Basile et~al.(2021)Basile, Fell, Fornaciari, Hovy, Paun, Plank, Poesio, and Uma}]{basile-etal-2021-need}
Valerio Basile, Michael Fell, Tommaso Fornaciari, Dirk Hovy, Silviu Paun, Barbara Plank, Massimo Poesio, and Alexandra Uma. 2021.
\newblock \href {https://aclanthology.org/2021.bppf-1.3} {{We Need to Consider Disagreement in Evaluation}}.
\newblock In \emph{{Proceedings of the 1st Workshop on Benchmarking: Past, Present and Future}}, pages 15--21, Online. Association for Computational Linguistics (ACL).

\bibitem[{Chan et~al.(2024)Chan, Jiayang, Wang, Jiang, Fang, Liu, and Song}]{chan-etal-2024-exploring}
Chunkit Chan, Cheng Jiayang, Weiqi Wang, Yuxin Jiang, Tianqing Fang, Xin Liu, and Yangqiu Song. 2024.
\newblock \href {https://aclanthology.org/2024.findings-eacl.47/} {{Exploring the Potential of ChatGPT on Sentence Level Relations: A Focus on Temporal, Causal, and Discourse Relations}}.
\newblock In \emph{{Findings of the 18th Conference of the European Chapter of the Association for Computational Linguistics (EACL'24)}}, pages 684--721, St. Julian's, Malta. Association for Computational Linguistics (ACL).

\bibitem[{Chan et~al.(2023)Chan, Liu, Cheng, Li, Song, Wong, and See}]{chan-etal-2023-DiscoPrompt}
Chunkit Chan, Xin Liu, Jiayang Cheng, Zihan Li, Yangqiu Song, Ginny~Y Wong, and Simon See. 2023.
\newblock \href {https://aclanthology.org/2023.findings-acl.4.pdf} {{DiscoPrompt: Path Prediction Prompt Tuning for Implicit Discourse Relation Recognition}}.
\newblock In \emph{{Findings of the 61st Annual Meeting of the Association for Computational Linguistics (ACL'23)}}, pages 35--57, Toronto, Ontario, Canada. Association for Computational Linguistics (ACL).

\bibitem[{Costa and Kosseim(2024)}]{costa2024exploring}
Nelson~Filipe Costa and Leila Kosseim. 2024.
\newblock \href {https://aclanthology.org/2024.codi-1.11} {{Exploring Soft-Label Training for Implicit Discourse Relation Recognition}}.
\newblock In \emph{{Proceedings of the 5th Workshop on Computational Approaches to Discourse (CODI'24)}}, pages 120--126, St. Julians, Malta. Association for Computational Linguistics (ACL).

\bibitem[{Costa and Kosseim(2025)}]{costa2025multi-lingual}
Nelson~Filipe Costa and Leila Kosseim. 2025.
\newblock {Multi-Lingual Implicit Discourse Relation Recognition with Multi-Label Hierarchical Learning}.
\newblock In \emph{{Proceedings of the 26th Annual Meeting of the Special Interest Group on Discourse and Dialogue (SIGDIAL'25)}}, Avignon, France. Association for Computational Linguistics (ACL).

\bibitem[{Costa et~al.(2023)Costa, Sheikh, and Kosseim}]{costa2023mapping}
Nelson~Filipe Costa, Nadia Sheikh, and Leila Kosseim. 2023.
\newblock \href {https://aclanthology.org/2023.ranlp-1.39} {{Mapping Explicit and Implicit Discourse Relations between the RST-DT and the PDTB 3.0}}.
\newblock In \emph{{Proceedings of the 14th International Conference on Recent Advances in Natural Language Processing (RANLP'23)}}, pages 344--352, Varna, Bulgaria.

\bibitem[{Demberg et~al.(2019)Demberg, Scholman, and Asr}]{demberg2019compatible}
Vera Demberg, Merel~CJ Scholman, and Fatemeh~Torabi Asr. 2019.
\newblock \href {https://doi.org/10.5087/dad.2019.104} {{How compatible are our discourse annotation frameworks? Insights from mapping RST-DT and PDTB annotations}}.
\newblock \emph{{Dialogue \& Discourse}}, 10(1):87--135.

\bibitem[{Devlin et~al.(2019)Devlin, Chang, Lee, and Toutanova}]{devlin-etal-2019-bert}
Jacob Devlin, Ming-Wei Chang, Kenton Lee, and Kristina Toutanova. 2019.
\newblock \href {https://aclanthology.org/N19-1423} {{BERT: Pre-training of Deep Bidirectional Transformers for Language Understanding}}.
\newblock In \emph{{Proceedings of the 2019 Conference of the North American Chapter of the Association for Computational Linguistics: Human Language Technologies (NAACL-HLT'19)}}, pages 4171--4186, Minneapolis, Minnesota, USA. Association for Computational Linguistics (ACL).

\bibitem[{Fornaciari et~al.(2021)Fornaciari, Uma, Paun, Plank, Hovy, and Poesio}]{fornaciari-etal-2021-beyond}
Tommaso Fornaciari, Alexandra Uma, Silviu Paun, Barbara Plank, Dirk Hovy, and Massimo Poesio. 2021.
\newblock \href {https://aclanthology.org/2021.naacl-main.204} {{Beyond Black {\&} White: Leveraging Annotator Disagreement via Soft-Label Multi-Task Learning}}.
\newblock In \emph{{Proceedings of the 2021 Conference of the North American Chapter of the Association for Computational Linguistics: Human Language Technologies (NAACL-HLT'21)}}, pages 2591--2597, Online. Association for Computational Linguistics (ACL).

\bibitem[{Grattafiori et~al.(2024)Grattafiori, Dubey, Jauhri, Pandey, Kadian, Al-Dahle, Letman, Mathur, Schelten, Vaughan et~al.}]{grattafiori2024llama}
Aaron Grattafiori, Abhimanyu Dubey, Abhinav Jauhri, Abhinav Pandey, Abhishek Kadian, Ahmad Al-Dahle, Aiesha Letman, Akhil Mathur, Alan Schelten, Alex Vaughan, and 1 others. 2024.
\newblock \href {https://arxiv.org/pdf/2407.21783} {{The Llama 3 Herd of Models}}.
\newblock \emph{arXiv preprint arXiv:2407.21783}.

\bibitem[{Hoek et~al.(2021)Hoek, Scholman, and Sanders}]{hoek-etal-2021-less}
Jet Hoek, Merel~C.J. Scholman, and Ted~J.M. Sanders. 2021.
\newblock \href {https://aclanthology.org/2021.discann-1.1} {{Is there less annotator agreement when the discourse relation is underspecified?}}
\newblock In \emph{{Proceedings of the 1st Workshop on Integrating Perspectives on Discourse Annotation}}, pages 1--6, Online. Association for Computational Linguistics (ACL).

\bibitem[{Ji and Eisenstein(2015)}]{ji-eisenstein-2015-one}
Yangfeng Ji and Jacob Eisenstein. 2015.
\newblock \href {https://aclanthology.org/Q15-1024} {{One Vector is Not Enough: Entity-Augmented Distributed Semantics for Discourse Relations}}.
\newblock \emph{{Transactions of the Association for Computational Linguistics (TACL)}}, 3:329--344.

\bibitem[{Jiang and de~Marneffe(2022)}]{jiang-marneffe-2022-investigating}
Nan-Jiang Jiang and Marie-Catherine de~Marneffe. 2022.
\newblock \href {https://aclanthology.org/2022.tacl-1.78} {{Investigating Reasons for Disagreement in Natural Language Inference}}.
\newblock \emph{{Transactions of the Association for Computational Linguistics (TACL)}}, 10:1357--1374.

\bibitem[{Kim et~al.(2020)Kim, Feng, Gunasekara, and Lastras}]{kim-etal-2020-implicit}
Najoung Kim, Song Feng, Chulaka Gunasekara, and Luis Lastras. 2020.
\newblock \href {https://aclanthology.org/2020.acl-main.480} {{Implicit Discourse Relation Classification: We Need to Talk about Evaluation}}.
\newblock In \emph{{Proceedings of the 58th Annual Meeting of the Association for Computational Linguistics (ACL'20)}}, pages 5404--5414, Online. Association for Computational Linguistics (ACL).

\bibitem[{Kingma and Ba(2015)}]{diederik2015adam}
Diederik~P. Kingma and Jimmy Ba. 2015.
\newblock \href {http://arxiv.org/abs/1412.6980} {{Adam: A Method for Stochastic Optimization}}.
\newblock In \emph{{Proceedings of the 3rd International Conference on Learning Representations (ICLR'15)}}, pages 1--15, San Diego, California, USA.

\bibitem[{Li et~al.(2024)Li, Yin, and Carenini}]{li-etal-2024-dialogue}
Chuyuan Li, Yuwei Yin, and Giuseppe Carenini. 2024.
\newblock \href {https://aclanthology.org/2024.SIGDIAL-1.1/} {{Dialogue Discourse Parsing as Generation: A Sequence-to-Sequence {LLM}-based Approach}}.
\newblock In \emph{{Proceedings of the 25th Annual Meeting of the Special Interest Group on Discourse and Dialogue (SIGDIAL'24)}}, pages 1--14, Kyoto, Japan. Association for Computational Linguistics (ACL).

\bibitem[{Lin(1991)}]{lin1991divergence}
Jianhua Lin. 1991.
\newblock {Divergence measures based on the Shannon entropy}.
\newblock \emph{{IEEE Transactions on Information Theory}}, 37(1):145--151.

\bibitem[{Lin et~al.(2009)Lin, Kan, and Ng}]{lin-etal-2009-recognizing}
Ziheng Lin, Min-Yen Kan, and Hwee~Tou Ng. 2009.
\newblock \href {https://aclanthology.org/D09-1036} {{Recognizing Implicit Discourse Relations in the Penn Discourse Treebank}}.
\newblock In \emph{{Proceedings of the 2009 Conference on Empirical Methods in Natural Language Processing (EMNLP'09)}}, pages 343--351, Singapore. Association for Computational Linguistics (ACL).

\bibitem[{Liu and Strube(2023)}]{liu-strube-2023-annotation}
Wei Liu and Michael Strube. 2023.
\newblock \href {https://aclanthology.org/2023.acl-long.874} {{Annotation-Inspired Implicit Discourse Relation Classification with Auxiliary Discourse Connective Generation}}.
\newblock In \emph{{Proceedings of the 61st Annual Meeting of the Association for Computational Linguistics (ACL'23)}}, pages 15696--15712, Toronto, Ontario, Canada. Association for Computational Linguistics (ACL).

\bibitem[{Liu et~al.(2019)Liu, Ott, Goyal, Du, Joshi, Chen, Levy, Lewis, Zettlemoyer, and Stoyanov}]{liu2019roberta}
Yinhan Liu, Myle Ott, Naman Goyal, Jingfei Du, Mandar Joshi, Danqi Chen, Omer Levy, Mike Lewis, Luke Zettlemoyer, and Veselin Stoyanov. 2019.
\newblock \href {https://arxiv.org/pdf/1907.11692.pdf} {{RoBERTa: A Robustly Optimized BERT Pretraining Approach}}.
\newblock \emph{arXiv preprint arXiv:1907.11692}.

\bibitem[{Long et~al.(2024)Long, Siddharth, and Webber}]{long2024multi}
Wanqiu Long, N~Siddharth, and Bonnie Webber. 2024.
\newblock \href {https://aclanthology.org/2024.findings-acl.500} {{Multi-Label Classification for Implicit Discourse Relation Recognition}}.
\newblock In \emph{Findings of the 62nd Annual Meeting of the Association for Computational Linguistics (ACL'24)}, pages 8437--8451, Bangkok, Thailand. Association for Computational Linguistics (ACL).

\bibitem[{Long and Webber(2022)}]{long-webber-2022-facilitating}
Wanqiu Long and Bonnie Webber. 2022.
\newblock \href {https://aclanthology.org/2022.emnlp-main.734} {{Facilitating Contrastive Learning of Discourse Relational Senses by Exploiting the Hierarchy of Sense Relations}}.
\newblock In \emph{{Proceedings of the 2022 Conference on Empirical Methods in Natural Language Processing (EMNLP'22)}}, pages 10704--10716, Abu Dhabi, United Arab Emirates. Association for Computational Linguistics (ACL).

\bibitem[{Long and Webber(2024)}]{long2024leveraging}
Wanqiu Long and Bonnie Webber. 2024.
\newblock \href {https://arxiv.org/pdf/2411.14880} {{Leveraging Hierarchical Prototypes as the Verbalizer for Implicit Discourse Relation Recognition}}.
\newblock \emph{arXiv preprint arXiv:2411.14880}.

\bibitem[{Loshchilov and Hutter(2017)}]{ilya2017sgdr}
Ilya Loshchilov and Frank Hutter. 2017.
\newblock \href {https://openreview.net/pdf/f1d82a22d77c21787940a33db6ce95a245c55eeb.pdf} {{SGDR: Stochastic Gradient Descent with Warm Restarts}}.
\newblock In \emph{{Proceedings of the 5th International Conference on Learning Representations (ICLR'17)}}, pages 1--16, Toulon, France.

\bibitem[{Miltsakaki et~al.(2004)Miltsakaki, Prasad, Joshi, and Webber}]{miltsakaki2004penn}
Eleni Miltsakaki, Rashmi Prasad, Aravind Joshi, and Bonnie Webber. 2004.
\newblock \href {http://www.lrec-conf.org/proceedings/lrec2004/pdf/618.pdf} {{The Penn Discourse Treebank}}.
\newblock In \emph{{Proceedings of the Fourth International Conference on Language Resources and Evaluation (LREC'04)}}, pages 2237--2240, Lisbon, Portugal. European Language Resources Association (ELRA).

\bibitem[{Nie et~al.(2020)Nie, Zhou, and Bansal}]{nie-etal-2020-learn}
Yixin Nie, Xiang Zhou, and Mohit Bansal. 2020.
\newblock \href {https://aclanthology.org/2020.emnlp-main.734} {{What Can We Learn from Collective Human Opinions on Natural Language Inference Data?}}
\newblock In \emph{{Proceedings of the 2020 Conference on Empirical Methods in Natural Language Processing (EMNLP'20)}}, pages 9131--9143, Online. Association for Computational Linguistics.

\bibitem[{Pavlick and Kwiatkowski(2019)}]{pavlick-kwiatkowski-2019-inherent}
Ellie Pavlick and Tom Kwiatkowski. 2019.
\newblock \href {https://aclanthology.org/Q19-1043} {{Inherent Disagreements in Human Textual Inferences}}.
\newblock \emph{{Transactions of the Association for Computational Linguistics (TACL)}}, 7:677--694.

\bibitem[{Plank(2022)}]{plank-2022-problem}
Barbara Plank. 2022.
\newblock \href {https://aclanthology.org/2022.emnlp-main.731} {{The {``}Problem{''} of Human Label Variation: On Ground Truth in Data, Modeling and Evaluation}}.
\newblock In \emph{{Proceedings of the 2022 Conference on Empirical Methods in Natural Language Processing (EMNLP'22)}}, pages 10671--10682, Abu Dhabi, United Arab Emirates. Association for Computational Linguistics (ACL).

\bibitem[{Prasad et~al.(2008{\natexlab{a}})Prasad, Dinesh, Lee, Miltsakaki, Robaldo, Joshi, and Webber}]{prasad2008penn}
Rashmi Prasad, Nikhil Dinesh, Alan Lee, Eleni Miltsakaki, Livio Robaldo, Aravind Joshi, and Bonnie Webber. 2008{\natexlab{a}}.
\newblock \href {http://www.lrec-conf.org/proceedings/lrec2008/pdf/754_paper.pdf} {{The Penn Discourse TreeBank 2.0.}}
\newblock In \emph{{Proceedings of the Sixth International Conference on Language Resources and Evaluation (LREC'08)}}, pages 2961--2968, Marrakech, Morocco. European Language Resources Association (ELRA).

\bibitem[{Prasad et~al.(2008{\natexlab{b}})Prasad, Lee, Dinesh, Miltsakaki, Campion, Joshi, and Webber}]{corpus-pdtb-2}
Rashmi Prasad, Alan Lee, Nikhil Dinesh, Eleni Miltsakaki, Geraud Campion, Aravind Joshi, and Bonnie Webber. 2008{\natexlab{b}}.
\newblock \href {https://doi.org/10.35111/nbvh-1n26} {{Penn Discourse Treebank Version 2.0}}.
\newblock LDC2008T05. Web Download. Philadelphia: Linguistic Data Consortium.

\bibitem[{Prasad et~al.(2019)Prasad, Webber, Lee, and Joshi}]{corpus-pdtb-3}
Rashmi Prasad, Bonnie Webber, Alan Lee, and Aravind Joshi. 2019.
\newblock \href {https://doi.org/10.35111/qebf-gk47} {{Penn Discourse Treebank Version 3.0}}.
\newblock LDC2019T05. Web Download. Philadelphia: Linguistic Data Consortium.

\bibitem[{Pyatkin et~al.(2020)Pyatkin, Klein, Tsarfaty, and Dagan}]{pyatkin-etal-2020-qadiscourse}
Valentina Pyatkin, Ayal Klein, Reut Tsarfaty, and Ido Dagan. 2020.
\newblock \href {https://aclanthology.org/2020.emnlp-main.224} {{QADiscourse - Discourse Relations as QA Pairs: Representation, Crowdsourcing and Baselines}}.
\newblock In \emph{{Proceedings of the 2020 Conference on Empirical Methods in Natural Language Processing (EMNLP'20)}}, pages 2804--2819, Online. Association for Computational Linguistics.

\bibitem[{Pyatkin et~al.(2023)Pyatkin, Yung, Scholman, Tsarfaty, Dagan, and Demberg}]{pyatkin2023design}
Valentina Pyatkin, Frances Yung, Merel C.~J. Scholman, Reut Tsarfaty, Ido Dagan, and Vera Demberg. 2023.
\newblock \href {https://doi.org/10.1162/tacl\_a\_00586} {{Design Choices for Crowdsourcing Implicit Discourse Relations: Revealing the Biases Introduced by Task Design}}.
\newblock \emph{Transactions of the Association for Computational Linguistics (TACL)}, 11:1014--1032.

\bibitem[{Raffel et~al.(2020)Raffel, Shazeer, Roberts, Lee, Narang, Matena, Zhou, Li, and Liu}]{raffel2020exploring}
Colin Raffel, Noam Shazeer, Adam Roberts, Katherine Lee, Sharan Narang, Michael Matena, Yanqi Zhou, Wei Li, and Peter~J Liu. 2020.
\newblock \href {https://jmlr.org/papers/volume21/20-074/20-074.pdf} {{Exploring the Limits of Transfer Learning with a Unified Text-to-Text Transformer}}.
\newblock \emph{{Journal of Machine Learning Research (JMLR)}}, 21(140):1--67.

\bibitem[{Rohde et~al.(2016)Rohde, Dickinson, Schneider, Clark, Louis, and Webber}]{rohde-etal-2016-filling}
Hannah Rohde, Anna Dickinson, Nathan Schneider, Christopher N.~L. Clark, Annie Louis, and Bonnie Webber. 2016.
\newblock \href {https://aclanthology.org/W16-1707} {{Filling in the Blanks in Understanding Discourse Adverbials: Consistency, Conflict, and Context-Dependence in a Crowdsourced Elicitation Task}}.
\newblock In \emph{{Proceedings of the 10th Linguistic Annotation Workshop (LAW'16)}}, pages 49--58, Berlin, Germany. Association for Computational Linguistics.

\bibitem[{Sanh et~al.(2019)Sanh, Debut, Chaumond, and Wolf}]{sanh2019distilbert}
Victor Sanh, Lysandre Debut, Julien Chaumond, and Thomas Wolf. 2019.
\newblock \href {https://arxiv.org/pdf/1910.01108} {{DistilBERT, a distilled version of BERT: smaller, faster, cheaper and lighter}}.
\newblock \emph{arXiv preprint arXiv:1910.01108}.

\bibitem[{Scholman and Demberg(2017)}]{scholman2017examples}
Merel Scholman and Vera Demberg. 2017.
\newblock \href {https://doi.org/10.5087/dad.2017.203} {{Examples and Specifications that Prove a Point: Identifying Elaborative and Argumentative Discourse Relations}}.
\newblock \emph{{Dialogue \& Discourse}}, 8(2):56--83.

\bibitem[{Scholman et~al.(2022{\natexlab{a}})Scholman, Dong, Yung, and Demberg}]{scholman-etal-2022-discogem}
Merel Scholman, Tianai Dong, Frances Yung, and Vera Demberg. 2022{\natexlab{a}}.
\newblock \href {https://aclanthology.org/2022.lrec-1.351} {{DiscoGeM: A Crowdsourced Corpus of Genre-Mixed Implicit Discourse Relations}}.
\newblock In \emph{{Proceedings of the 13th Language Resources and Evaluation Conference (LREC'22)}}, pages 3281--3290, Marseille, France. European Language Resources Association (ELRA).

\bibitem[{Scholman et~al.(2022{\natexlab{b}})Scholman, Pyatkin, Yung, Dagan, Tsarfaty, and Demberg}]{scholman-etal-2022-design}
Merel Scholman, Valentina Pyatkin, Frances Yung, Ido Dagan, Reut Tsarfaty, and Vera Demberg. 2022{\natexlab{b}}.
\newblock \href {https://aclanthology.org/2022.lrec-1.231} {{Design Choices in Crowdsourcing Discourse Relation Annotations: The Effect of Worker Selection and Training}}.
\newblock In \emph{{Proceedings of the 13th Language Resources and Evaluation Conference (LREC'22)}}, pages 2148--2156, Marseille, France. European Language Resources Association (ELRA).

\bibitem[{Shi and Demberg(2017)}]{shi-demberg-2017-need}
Wei Shi and Vera Demberg. 2017.
\newblock \href {https://aclanthology.org/E17-2024} {{On the Need of Cross Validation for Discourse Relation Classification}}.
\newblock In \emph{{Proceedings of the 15th Conference of the European Chapter of the Association for Computational Linguistics (EACL'17)}}, pages 150--156, Valencia, Spain. Association for Computational Linguistics (ACL).

\bibitem[{Shi and Demberg(2019)}]{shi-demberg-2019-next}
Wei Shi and Vera Demberg. 2019.
\newblock \href {https://aclanthology.org/D19-1586} {{Next Sentence Prediction helps Implicit Discourse Relation Classification within and across Domains}}.
\newblock In \emph{{Proceedings of the 2019 Conference on Empirical Methods in Natural Language Processing and the 9th International Joint Conference on Natural Language Processing (EMNLP-IJCNLP'19)}}, pages 5790--5796, Hong Kong, China. Association for Computational Linguistics (ACL).

\bibitem[{Stede(2008)}]{stede2008disambiguation}
Manfred Stede. 2008.
\newblock \href {https://doi.org/10.1007/s11168-008-9053-7} {{Disambiguating Rhetorical Structure}}.
\newblock \emph{{Research on Language and Computation}}, 6(3):311--332.

\bibitem[{Thompson et~al.(2024)Thompson, Chaturvedi, Hunter, and Asher}]{thompson-etal-2024-llamipa}
Kate Thompson, Akshay Chaturvedi, Julie Hunter, and Nicholas Asher. 2024.
\newblock \href {https://aclanthology.org/2024.findings-emnlp.373/} {{Llamipa: An Incremental Discourse Parser}}.
\newblock In \emph{{Findings of 2024 Conference on Empirical Methods in Natural Language Processing (EMNLP'24)}}, pages 6418--6430, Miami, Florida, USA. Association for Computational Linguistics (ACL).

\bibitem[{Uma et~al.(2021)Uma, Fornaciari, Hovy, Paun, Plank, and Poesio}]{uma2021learning}
Alexandra~N Uma, Tommaso Fornaciari, Dirk Hovy, Silviu Paun, Barbara Plank, and Massimo Poesio. 2021.
\newblock \href {https://doi.org/10.1613/jair.1.12752} {{Learning from Disagreement: A Survey}}.
\newblock \emph{{Journal of Artificial Intelligence Research}}, 72(1):1385--1470.

\bibitem[{van~der Meer et~al.(2024)van~der Meer, Falk, Murukannaiah, and Liscio}]{van2024annotator}
Michiel van~der Meer, Neele Falk, Pradeep~K Murukannaiah, and Enrico Liscio. 2024.
\newblock \href {https://aclanthology.org/2024.emnlp-main.1031/} {{Annotator-Centric Active Learning for Subjective NLP Tasks}}.
\newblock In \emph{{Proceedings of the 2024 Conference on Empirical Methods in Natural Language Processing (EMNLP'24)}}, pages 18537--18555, Miami, Florida, USA. Association for Computational Linguistics (ACL).

\bibitem[{Webber et~al.(2019)Webber, Prasad, Lee, and Joshi}]{webber2019penn}
Bonnie Webber, Rashmi Prasad, Alan Lee, and Aravind Joshi. 2019.
\newblock \href {https://catalog.ldc.upenn.edu/docs/LDC2019T05/PDTB3-Annotation-Manual.pdf} {{The Penn Discourse Treebank 3.0 Annotation Manual}}.
\newblock Technical report, University of Pennsylvania.

\bibitem[{Yung et~al.(2024)Yung, Ahmad, Scholman, and Demberg}]{yung-etal-2024-prompting}
Frances Yung, Mansoor Ahmad, Merel Scholman, and Vera Demberg. 2024.
\newblock \href {https://aclanthology.org/2024.law-1.15/} {{Prompting Implicit Discourse Relation Annotation}}.
\newblock In \emph{{Proceedings of the 18th Linguistic Annotation Workshop (LAW-XVIII)}}, pages 150--165, St. Julian's, Malta. Association for Computational Linguistics (ACL).

\bibitem[{Yung et~al.(2022)Yung, Anuranjana, Scholman, and Demberg}]{yung-etal-2022-label}
Frances Yung, Kaveri Anuranjana, Merel Scholman, and Vera Demberg. 2022.
\newblock \href {https://aclanthology.org/2022.codi-1.7} {{Label distributions help implicit discourse relation classification}}.
\newblock In \emph{{Proceedings of the 3rd Workshop on Computational Approaches to Discourse (CODI'22)}}, pages 48--53, Gyeongju, Republic of Korea. International Conference on Computational Linguistics (ICCL).

\bibitem[{Yung and Demberg(2025)}]{yung-demberg-2025-crowdsourcing}
Frances Yung and Vera Demberg. 2025.
\newblock \href {https://aclanthology.org/2025.comedi-1.2/} {{On Crowdsourcing Task Design for Discourse Relation Annotation}}.
\newblock In \emph{{Proceedings of Context and Meaning: Navigating Disagreements in NLP Annotation}}, pages 12--19, Abu Dhabi, UAE. International Committee for Computational Linguistics (ICCL).

\bibitem[{Yung et~al.(2019)Yung, Demberg, and Scholman}]{yung-etal-2019-crowdsourcing}
Frances Yung, Vera Demberg, and Merel Scholman. 2019.
\newblock \href {https://aclanthology.org/W19-4003} {{Crowdsourcing Discourse Relation Annotations by a Two-Step Connective Insertion Task}}.
\newblock In \emph{{Proceedings of the 13th Linguistic Annotation Workshop (LAW'19)}}, pages 16--25, Florence, Italy. Association for Computational Linguistics (ACL).

\bibitem[{Zeng et~al.(2024)Zeng, He, Sun, Xu, Liu, and Wang}]{zeng-etal-2024-global}
Lei Zeng, Ruifang He, Haowen Sun, Jing Xu, Chang Liu, and Bo~Wang. 2024.
\newblock \href {https://aclanthology.org/2024.lrec-main.686/} {{Global and Local Hierarchical Prompt Tuning Framework for Multi-level Implicit Discourse Relation Recognition}}.
\newblock In \emph{{Proceedings of the 2024 Joint International Conference on Computational Linguistics, Language Resources and Evaluation (LREC-COLING'24)}}, pages 7760--7773, Torino, Italia. European Language Resources Association (ELRA) and International Committee for Computational Linguistics (ICCL).

\bibitem[{Zhao et~al.(2023)Zhao, He, Xiao, and Xu}]{zhao-etal-2023-infusing}
Haodong Zhao, Ruifang He, Mengnan Xiao, and Jing Xu. 2023.
\newblock \href {https://aclanthology.org/2023.acl-long.357} {{Infusing Hierarchical Guidance into Prompt Tuning: A Parameter-Efficient Framework for Multi-level Implicit Discourse Relation Recognition}}.
\newblock In \emph{{Proceedings of the 61st Annual Meeting of the Association for Computational Linguistics (ACL'23)}}, pages 6477--6492, Toronto, Ontario, Canada. Association for Computational Linguistics (ACL).

\end{thebibliography}


\appendix

\section{Data Statistics}
\label{apx:statistics}

\begin{figure*}
    \centering
    \includegraphics[width=\textwidth]{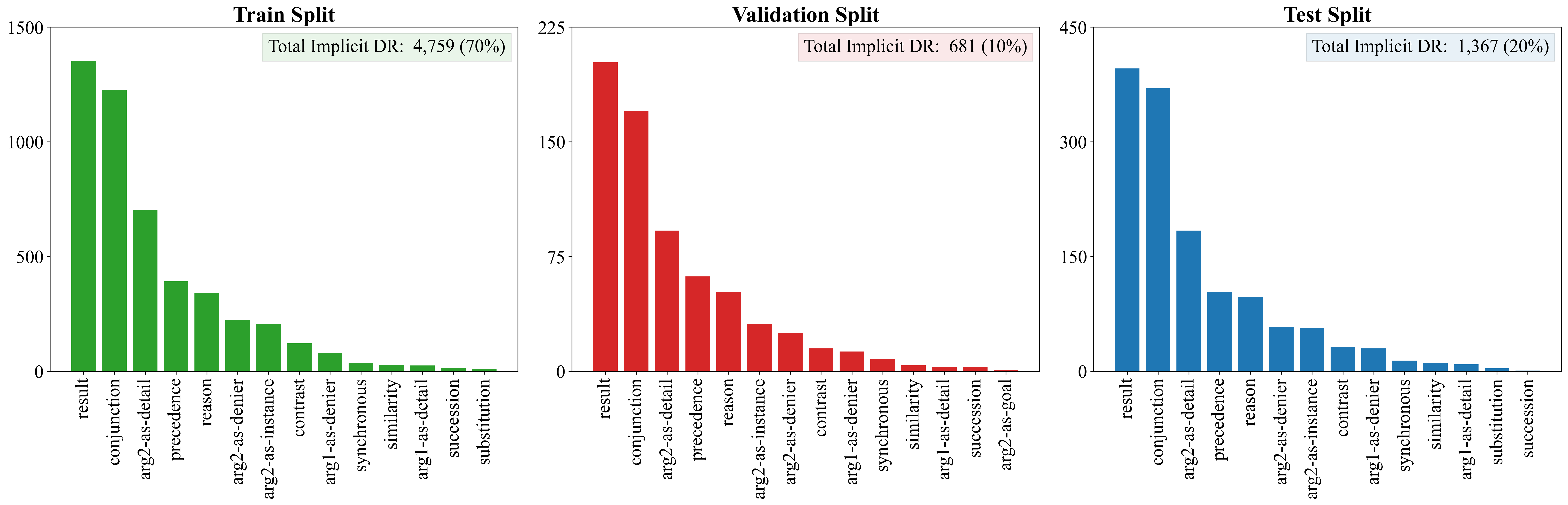}
    \caption{Distribution of majority label senses across all splits of the DiscoGeM corpus (see Section~\ref{sec:disco-data}).}
    \label{fig:discogem-label-distribution}
\end{figure*}

To standardize and ensure a fair comparison of results on IDRR, \citet{kim-etal-2020-implicit} proposed a set of 14 level-2 senses in the PDTB~3.0 framework. However, the \textsc{Cause+Belief} sense in this set was not annotated in DiscoGeM. Therefore, we adapted the standard set to accommodate the existing level-2 senses in the corpus by replacing \textsc{Cause+Belief} with \textsc{Similarity}. We then removed all senses not included in the standard set from the corpus and normalized the distribution values over the remaining senses following the L1 norm. This ensures the sum of each distribution adds to 1 for each instance while preserving the relative distance between senses. Table~\ref{tab:multi-statistics} shows the distribution of senses across all levels in the original DiscoGeM corpus and in our adapted set of 14 level-2 senses. Each value represents the sum of the corresponding sense across all instances. We then split 70\% of the corpus for training, 10\% for validation and 20\% for testing. To ensure a balanced distribution, we kept the same distribution of majority labels across all data splits as shown in Figure~\ref{fig:discogem-label-distribution} in Section \ref{sec:disco-data}. In the single-label setting, we replaced the full multi-label sense distribution of each discourse relation by its majority sense. Table~\ref{tab:single-statistics-test} shows the level-2 majority senses in the test set of DiscoGeM and reference senses in the different test splits of the PDTB~3.0 (see Section~\ref{sec:pdtb-data}).

\begin{table*}
    \centering
    \renewcommand{\arraystretch}{1.2}
    \resizebox{\textwidth}{!}{
        \begin{tabular}{|c|c|c|c|c|c|c|c|c|}
            \hline
            \hline
            \multirow{2}{*}{Level-1} & \multicolumn{2}{c|}{Sum} & \multirow{2}{*}{Level-2} & \multicolumn{2}{c|}{Sum} & \multirow{2}{*}{Level-3} & \multicolumn{2}{c|}{Sum} \\
            \cline{2-3} \cline{5-6} \cline{8-9}
             & \multicolumn{1}{c|}{Original} & \multicolumn{1}{c|}{Adapted} &  & \multicolumn{1}{c|}{Original} & \multicolumn{1}{c|}{Adapted} &  & \multicolumn{1}{c|}{Original} & \multicolumn{1}{c|}{Adapted} \\
            \hline
            \hline
            \multirow{3}{*}{\textsc{Temporal}} & \multirow{3}{*}{$584.3$} & \multirow{3}{*}{$619.5$} & \textsc{Synchronous} & $95.8$ & $102.7$ & - & - & - \\
            \cline{4-9}
             &  &  & \multirow{2}{*}{\textsc{Asynchronous}} & \multirow{2}{*}{$488.5$} & \multirow{2}{*}{$516.8$} & \textsc{Precedence} & $448.1$ & $474.2$ \\
             &  &  &  &  &  & \textsc{Succession} & $40.5$ & $42.7$ \\
            \hline
            \multirow{8}{*}{\textsc{Contingency}} & \multirow{8}{*}{$1,745.1$} & \multirow{8}{*}{$1,822.9$} & \multirow{3}{*}{\textsc{Cause}} & \multirow{3}{*}{$1,740.2$} & \multirow{3}{*}{$1,819.0$} & \textsc{Reason} & $382.5$ & $400.1$ \\
             &  &  &  &  &  & \textsc{Result} & $1,357.7$ & $1,418.9$ \\
             &  &  &  &  &  & \textsc{NegResult} & $0.0$ & $0.0$ \\
            \cline{4-9}
             &  &  & \multirow{2}{*}{\textsc{Condition}} & \multirow{2}{*}{$1.2$} & \multirow{2}{*}{$1.2$} & \textsc{ARG1-as-Cond} & $0.1$ & $0.1$ \\
             &  &  &  &  &  & \textsc{ARG2-as-Cond} & $1.1$ & $1.1$ \\
            \cline{4-9}
             &  &  & \multirow{2}{*}{\textsc{Neg-Condition}} & \multirow{2}{*}{$1.1$} & \multirow{2}{*}{$0.0$} & \textsc{ARG1-as-NegCond} & $1.0$ & $0.0$ \\
             &  &  &  &  &  & \textsc{ARG2-as-NegCond} & $0.1$ & $0.0$ \\
            \cline{4-9}
             &  &  & \multirow{2}{*}{\textsc{Purpose}} & \multirow{2}{*}{$2.6$} & \multirow{2}{*}{$2.7$} & \textsc{ARG1-as-Goal} & $1.7$ & $1.7$ \\
             &  &  &  &  &  & \textsc{ARG2-as-Goal} & $0.9$ & $0.9$ \\
            \hline
            \multirow{4}{*}{\textsc{Comparison}} & \multirow{4}{*}{$831.5$} & \multirow{4}{*}{$878.0$} & \multirow{2}{*}{\textsc{Concession}} & \multirow{2}{*}{$517.1$} & \multirow{2}{*}{$548.7$} & \textsc{ARG1-as-Denier} & $169.3$ & $179.9$ \\
             &  &  &  &  &  & \textsc{ARG2-as-Denier} & $347.7$ & $368.8$ \\
            \cline{4-9}
             &  &  & \textsc{Contrast} & $202.4$ & $213.4$ & - & - & - \\
            \cline{4-9}
             &  &  & \textsc{Similarity} & $112.0$ & $116.0$ & - & - & - \\
            \hline
            \multirow{13}{*}{\textsc{Expansion}} & \multirow{13}{*}{$3,034.6$} & \multirow{13}{*}{$3,184.6$} & \textsc{Conjunction} & $1,441.1$ & $1,518.0$ & - & - & - \\
            \cline{4-9}
             &  &  & \textsc{Disjunction} & $2.9$ & $0.0$ & - & - & - \\
            \cline{4-9}
             &  &  & \textsc{Equivalence} & $19.0$ & $19.8$ & - & - & - \\
            \cline{4-9}
             &  &  & \multirow{2}{*}{\textsc{Exception}} & \multirow{2}{*}{$1.4$} & \multirow{2}{*}{$0.0$} & \textsc{ARG1-as-Exception} & $0.3$ & $0.0$ \\
             &  &  &  &  &  & \textsc{ARG2-as-Exception} & $1.3$ & $0.0$ \\
            \cline{4-9}
             &  &  & \multirow{2}{*}{\textsc{Instantiation}} & \multirow{2}{*}{$388.8$} & \multirow{2}{*}{$403.8$} & \textsc{ARG1-as-Instance} & $16.9$ & $17.8$ \\
             &  &  &  &  &  & \textsc{ARG2-as-Instance} & $371.9$ & $386.0$ \\
            \cline{4-9}
             &  &  & \multirow{2}{*}{\textsc{Level-of-Detail}} & \multirow{2}{*}{$1,137.0$} & \multirow{2}{*}{$1,196.1$} & \textsc{ARG1-as-Detail} & $160.8$ & $170.2$ \\
             &  &  &  &  &  & \textsc{ARG2-as-Detail} & $976.2$ & $1,025.9$ \\
            \cline{4-9}
             &  &  & \multirow{2}{*}{\textsc{Manner}} & \multirow{2}{*}{$4.6$} & \multirow{2}{*}{$4.8$} & \textsc{ARG1-as-Manner} & $1.3$ & $1.3$ \\
             &  &  &  &  &  & \textsc{ARG2-as-Manner} & $3.3$ & $3.4$ \\
            \cline{4-9}
             &  &  & \multirow{2}{*}{\textsc{Substitution}} & \multirow{2}{*}{$39.7$} & \multirow{2}{*}{$42.1$} & \textsc{ARG1-as-Substitution} & $0.0$ & $0.0$ \\
             &  &  &  &  &  & \textsc{ARG2-as-Substitution} & $39.7$ & $42.1$ \\
            \hline
            \hline
        \end{tabular}
    }
    \caption{Distribution of senses across all levels in the original DiscoGeM and in our adapted set of 14 level-2 senses (see Section~\ref{sec:disco-data}). Each value represents the sum of the corresponding sense across all instances.}
    \label{tab:multi-statistics}
\end{table*}

\begin{table*}
    \centering
    \renewcommand{\arraystretch}{1.2}
    \scalebox{1}{
        \begin{tabular}{|c|c|c|c|c|c|}
            \hline
            \hline
            \multirow{2}{*}{Level-1} & \multirow{2}{*}{Level-2} & DiscoGeM & \multicolumn{3}{c|}{PDTB~3.0} \\
            \cline{3-6}
             &  & \multicolumn{1}{c|}{Test} & \multicolumn{1}{c|}{Lin} & \multicolumn{1}{c|}{Ji} & \multicolumn{1}{c|}{Cross} \\
            \hline
            \hline
            \multirow{2}{*}{\textsc{Temporal}} & \textsc{Synchronous} & $13$ & $19$ & $43$ & $42.85 \pm 11.94$ \\
            \cline{2-6}
             & \textsc{Asynchronous} & $103$ & $33$ & $105$ & $99.54 \pm 24.95$ \\
            \hline
            \multirow{3}{*}{\textsc{Contingency}} & \textsc{Cause} & $515$ & $271$ & $406$ & $465.38 \pm 77.28$ \\
            \cline{2-6}
             & \textsc{Condition} & $0$ & $10$ & $15$ & $16.31 \pm 3.61$ \\
            \cline{2-6}
             & \textsc{Purpose} & $0$ & $46$ & $89$ & $110.08 \pm 22.60$ \\
            \hline
            \multirow{3}{*}{\textsc{Comparison}} & \textsc{Concession} & $104$ & $83$ & $98$ & $121.23 \pm 22.87$ \\
            \cline{2-6}
             & \textsc{Contrast} & $27$ & $52$ & $54$ & $69.92 \pm 18.99$ \\
            \cline{2-6}
             & \textsc{Similarity} & $10$ & $1$ & $2$ & $2.23 \pm 1.48$ \\
            \hline
            \multirow{6}{*}{\textsc{Expansion}} & \textsc{Conjunction} & $336$ & $161$ & $236$ & $349.38 \pm 86.25$ \\
            \cline{2-6}
             & \textsc{Equivalence} & $0$ & $24$ & $30$ & $27.54 \pm 10.27$ \\
            \cline{2-6}
             & \textsc{Instantiation} & $53$ & $70$ & $124$ & $118.62 \pm 24.97$ \\
            \cline{2-6}
             & \textsc{Level-of-Detail} & $202$ & $194$ & $208$ & $263.69 \pm 26.94$ \\
            \cline{2-6}
             & \textsc{Manner} & $0$ & $4$ & $17$ & $17.92 \pm 6.49$ \\
            \cline{2-6}
             & \textsc{Substitution} & $4$ & $27$ & $26$ & $30.23 \pm 8.28$ \\
            \hline
            \hline
            \multicolumn{2}{|c|}{Total} & $1,367$ & $995$ & $1,453$ & $1,734.92 \pm 271.89$ \\
            \hline
            \hline
        \end{tabular}
    }
    \caption{Distribution of level-2 majority label senses across the test set of DiscoGeM (see Section~\ref{sec:disco-data}) and the different test splits of the PDTB~3.0 (see Section~\ref{sec:pdtb-data}). The values in PDTB~3.0 Cross correspond to the averaged distribution of majority label senses across all 12 test folds.}
    \label{tab:single-statistics-test}
\end{table*}


\section{Loss Functions}
\label{apx:loss}

We evaluated the performance of different loss functions on multi- and single-label IDRR. Considering $\hat{y}^h$ as the predicted distribution of the classification head $h$ (with $h \in \{1, 2, 3\}$), $y^h$ as the corresponding target distribution, $C^h$ the number of senses at the sense level-$h$ and $N$ the number of instances in each batch, we evaluated the following loss functions:

\begin{itemize}
    \item \underline{Cross-Entropy (CE)}

    \begin{equation}
        \scalebox{0.8}{$loss^h(\hat{y}^h,y^h) = -\frac{1}{N}\sum^N_{n=1}\sum^{C^h}_{c=1}log\Biggl(\frac{e^{\hat{y}^h_{n,c}}}{\sum^{C^h}_{i=1}e^{\hat{y}^h_{n,i}}}\Biggr)y^h_{n,c}$}
        \label{eq:single-loss}
    \end{equation}

\end{itemize}

\begin{itemize}
    \item \underline{Mean Absolute Error (MAE)}

    \begin{equation}
        \scalebox{0.9}{$loss^h(\hat{y}^h,y^h) = \frac{1}{NC^h}\sum^N_{n=1}\sum^{C^h}_{i=1} |\hat{y}^h_{n,i}-y^h_{n,i}|
        \label{eq:multi-loss}$}
    \end{equation}
\end{itemize}

\begin{itemize}
    \item \underline{Mean Squared Error (MSE)}

    \begin{equation}
    \scalebox{0.9}{$loss^h(\hat{y}^h,y^h) = \frac{1}{NC^h}\sum^N_{n=1}\sum^{C^h}_{i=1} (\hat{y}^h_{n,i}-y^h_{n,i})^2$}
        \label{eq:mse}
    \end{equation}
\end{itemize}

\begin{itemize}
    \item \underline{Huber Loss (Huber)}

    \begin{equation}
        \scalebox{0.8}{$loss^k(\hat{y}^h,y^h) = \sum^N_{n=1}\sum^{C^h}_{i=1}
        \begin{cases}
            \frac{{\delta^h_{n,i}}^2}{2NC^h}   & \text{, if } |\delta^h_{n,i}| < 1 \\[0.4cm]
            \frac{2|\delta^h_{n,i}|-1}{2NC^h}   & \text{, otherwise}
        \end{cases}$}
        \label{eq:huber}
    \end{equation}
    
    with
    
    \begin{equation}
        \delta^h_{n,i} = \hat{y}^h_{n,i}-y^h_{n,i}
        \label{eq:delta}
    \end{equation}
\end{itemize}

\begin{table*}
    \centering
    \renewcommand{\arraystretch}{1.2}
    \resizebox{\textwidth}{!}{
        \begin{tabular}{|c|c|ccc|ccc|}
            \hline
            \hline
            \multirow{2}{*}{Model} & \multirow{2}{*}{Loss} & \multicolumn{3}{c|}{JS Distance \scalebox{0.9}{$\searrow$} (Multi-Label)} & \multicolumn{3}{c|}{F1-Score \scalebox{0.9}{$\nearrow$} (Single-Label)} \\
            \cline{3-8}
             &  & \multicolumn{1}{c}{Level-1} & \multicolumn{1}{c}{Level-2} & \multicolumn{1}{c|}{Level-3} & \multicolumn{1}{c}{Level-1} & \multicolumn{1}{c}{Level-2} & \multicolumn{1}{c|}{Level-3} \\
            \hline
            \hline
            \multirow{2}{*}{BERT} & MSE & $0.319 \pm 0.005$ & $0.462 \pm 0.002$ & $0.544 \pm 0.005$ & $63.82 \pm 0.83$ & $48.68 \pm 0.16$ & $41.36 \pm 1.55$ \\
             & Huber & $0.318 \pm 0.007$ & $0.463 \pm 0.001$ & $0.542 \pm 0.002$ & $64.36 \pm 0.48$ & $49.45 \pm 0.96$ & $41.58 \pm 1.08$ \\
            \hline
            \multirow{2}{*}{DistilBERT} & MSE & $0.336 \pm 0.007$ & $0.481 \pm 0.003$ & $0.558 \pm 0.003$ & $60.10 \pm 0.96$ & $43.77 \pm 0.86$ & $37.66 \pm 1.14$ \\
             & Huber & $0.337 \pm 0.002$ & $0.479 \pm 0.003$ & $0.557 \pm 0.006$ & $60.41 \pm 0.86$ & $44.06 \pm 0.64$ & $37.29 \pm 0.28$ \\
            \hline
            \multirow{2}{*}{RoBERTa} & MSE & $0.308 \pm 0.012$ & $0.455 \pm 0.001$ & \bm{$0.534 \pm 0.005$} & \bm{$65.28 \pm 0.36$} & \bm{$53.94 \pm 1.80$} & \bm{$44.91 \pm 1.31$} \\
             & Huber & \bm{$0.305 \pm 0.007$} & \bm{$0.450 \pm 0.004$} & $0.539 \pm 0.003$ & $65.13 \pm 0.60$ & $53.39 \pm 0.19$ & $44.33 \pm 0.64$ \\
            \hline
            \multirow{2}{*}{DistilRoBERTa} & MSE & $0.313 \pm 0.004$ & $0.470 \pm 0.005$ & $0.555 \pm 0.003$ & $64.32 \pm 1.99$ & $50.48 \pm 0.29$ & $42.43 \pm 0.22$ \\
             & Huber & $0.318 \pm 0.004$ & $0.472 \pm 0.001$ & $0.561 \pm 0.004$ & $64.11 \pm 0.72$ & $50.46 \pm 0.73$ & $42.41 \pm 0.30$ \\
            \hline
            \hline
        \end{tabular}
    }
    \caption{Results of experimenting with different pre-trained language models and different loss functions in multi-label classification (JS distance) and in single-label classification (weighted F1-score). The results were averaged across three different runs with random starts. Values in bold show the best score for each metric.}
    \label{tab:model-selection-2}
\end{table*}

The MAE and MSE losses in Equations~\ref{eq:multi-loss}~and~\ref{eq:mse}, respectively, aim to minimize the overall difference between the predicted and target distributions by penalizing errors across all possible labels. MAE minimizes the average absolute differences, leading to predictions that are closer to the target distribution in an averaged sense, while MSE places a larger penalty on larger errors due to its quadratic nature, which can result in a stronger emphasis on outliers. The Huber loss in Equation~\ref{eq:huber} combines the properties of MAE and MSE in function of the delta parameter defined in Equation~\ref{eq:delta}. In other words, the Huber loss behaves like MAE for smaller errors and like MSE for larger errors. The results of experimenting with the MAE, MSE and Huber losses were therefore relatively similar. However, the MAE was slightly better and thus better suited for multi-label IDRR as shown in Table~\ref{tab:model-selection-1} in Section~\ref{sec:model-selection}. Table~\ref{tab:model-selection-2} shows the results of experimenting with the MSE and Huber loss functions in multi-label IDRR (JS distance) and in single-label IDRR (weighted F1-score). Conversely, the CE loss in Equation~\ref{eq:single-loss} focuses on maximizing the probability of the label with the highest score in the target distribution which makes it better suited for single-label IDRR as shown in Table~\ref{tab:model-selection-1} in Section~\ref{sec:model-selection}.

\section{Confusion Matrices}
\label{apx:confusion}

\begin{figure}
    \centering
    \includegraphics[width=\columnwidth]{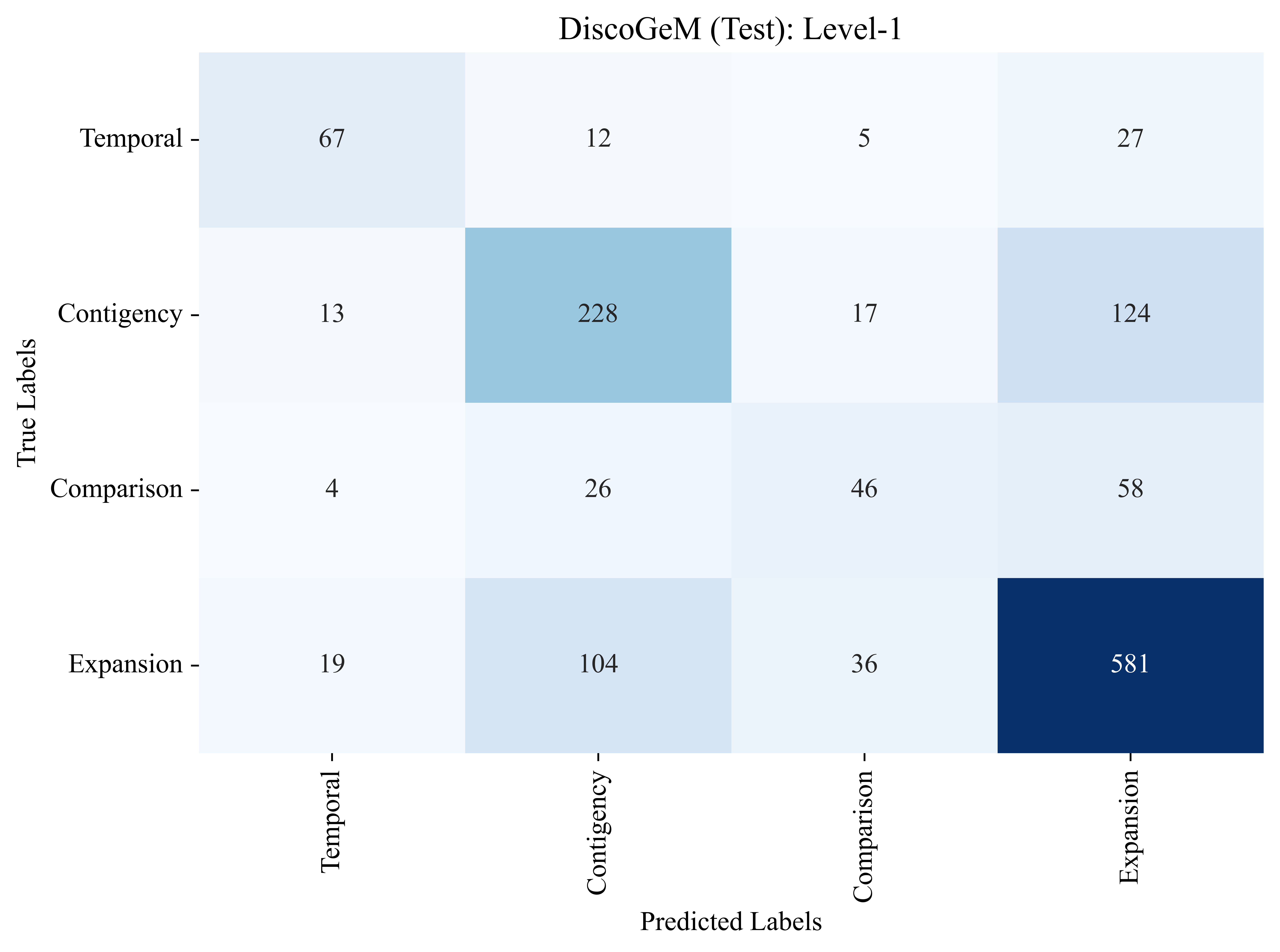}
    \caption{Confusion matrix for the individual per-class results of each sense in level-1 on the test set of DiscoGeM.}
    \label{fig:confusion_l1}
\end{figure}

\begin{figure}
    \centering
    \includegraphics[width=\columnwidth]{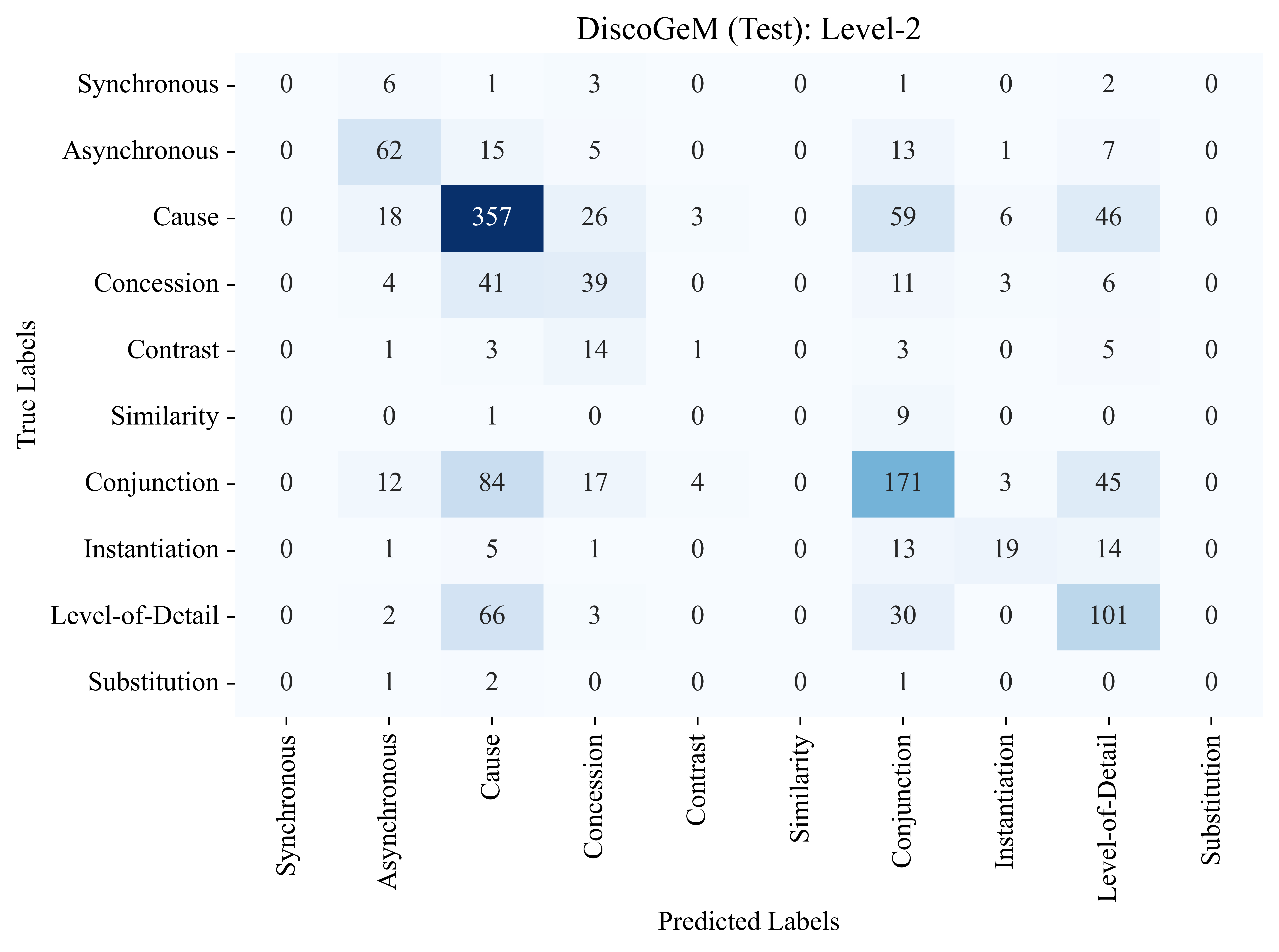}
    \caption{Confusion matrix for the individual per-class results of each sense in level-2 on the test set of DiscoGeM.}
    \label{fig:confusion_l2}
\end{figure}

To provide further details into the per-sense performance on the test set of DiscoGeM in Section~\ref{sec:individual}, we generated a confusion matrix at level-1 and level-2 for the results of a single run on the test set of DiscoGeM as shown in Figures~\ref{fig:confusion_l1}~and~\ref{fig:confusion_l2}, respectively. The confusion matrix in Figure~\ref{fig:confusion_l1} shows that, with the exception with \textsc{Comparison}, the most predicted label always aligns with the correct label. Nevertheless, the confusion matrix also shows that the model is more biased towards the most represented senses level-1 senses \textsc{Expansion} and \textsc{Contingency} (see Table~\ref{tab:multi-statistics} in Appendix~\ref{apx:statistics}). At the level-2, the confusion matrix in Table~\ref{fig:confusion_l2} shows that the model is not able to predict less represented senses in DiscoGeM (see Table~\ref{tab:multi-statistics} in Appendix~\ref{apx:statistics}), such as \textsc{Synchronous}, \textsc{Similarity} and \textsc{Substitution}. However, for the other senses, the most predicted label often aligns with the correct label.

\end{document}